\documentclass{article}

\usepackage[preprint]{neurips_2026}

\usepackage[utf8]{inputenc}
\usepackage[T1]{fontenc}
\usepackage{hyperref}
\usepackage{url}
\usepackage{booktabs}
\usepackage{array}
\usepackage{amsfonts}
\usepackage{amsmath}
\usepackage{nicefrac}
\usepackage{microtype}
\usepackage{pifont}
\usepackage{tabularx}
\usepackage{xcolor}
\usepackage{graphicx}

\newcommand{\cmark}{\ding{51}}
\newcommand{\xmark}{\ding{55}}

\title{\textsc{BigFinanceBench}: A Workflow-Grounded Benchmark for Financial-Research Agents}

\author{
  \textbf{Alex Wang}\textsuperscript{1,*} \quad
  \textbf{Georg Meinhardt}\textsuperscript{1,*} \quad
  \textbf{Jacob Katz}\textsuperscript{1} \\
  \textbf{Joseph H. Kim}\textsuperscript{2,\textdagger} \quad
  \textbf{Pratyush K. Chaudhary}\textsuperscript{1} \quad
  \textbf{Chase Blagden}\textsuperscript{2,\textdagger} \quad
  \textbf{Eric Xu}\textsuperscript{1} \\
  \textsuperscript{1}Rogo \quad
  \textsuperscript{2}OpenAI \\
  \texttt{alexwang@rogo.ai, georg@rogo.ai} \\
}

\begin{document}

\maketitle
\begingroup
\renewcommand{\thefootnote}{\fnsymbol{footnote}}
\footnotetext[1]{Equal contribution.}
\footnotetext[2]{Work done while at Rogo.}
\endgroup

\begin{abstract}
Financial-research answers are decision-relevant only when another analyst can audit how they
were produced: which source was chosen, which period and accounting definition were used, which
assumptions were made, and how the calculation was performed.
Existing finance benchmarks largely evaluate isolated subskills or final answers, leaving the
auditable derivation itself under-measured.
We introduce \textsc{BigFinanceBench}, a $928$-item expert-authored benchmark of open-ended
financial-research tasks in which each item pairs a ground-truth reference answer with a
point-weighted rubric that decomposes the derivation into independently checkable steps.
\textsc{BigFinanceBench} is \emph{workflow-grounded} in that it rates the whole derivation,
rather than just the final output.
Across $36{,}241$ rubric points, the benchmark supports partial-credit evaluation and
localization of failures across the analyst workflow.
Evaluating ten current frontier and open-weight agents, we find substantial headroom: the best
system reaches only $58.8\%$ rubric score, final-answer accuracy is a useful but lossy proxy
for derivation quality, and model capability varies non-uniformly across financial workflows.

\end{abstract}

\begin{figure}[!h]
  \centering
  \includegraphics[width=\linewidth]{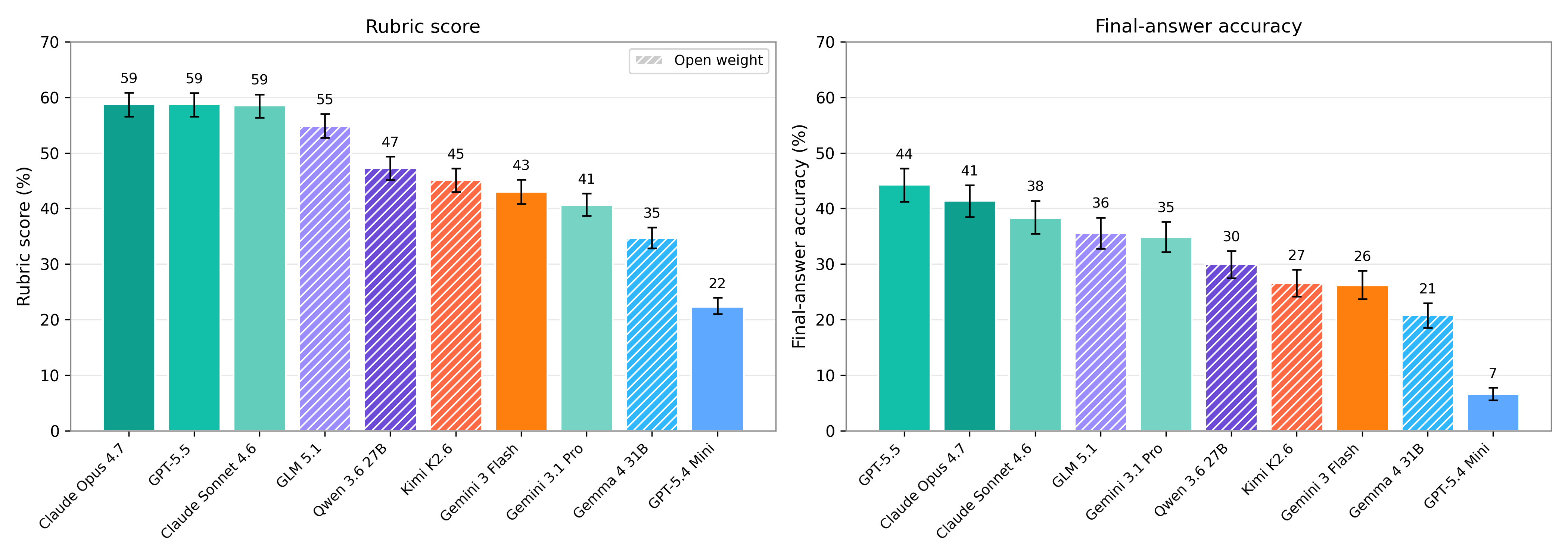}
  \caption{Model performance on \textsc{BigFinanceBench}: point-weighted rubric score
  (left) and final-answer accuracy (right). Error bars show $95\%$ bootstrap confidence
  intervals over questions; hatched bars are open-weight systems.}
  \label{fig:motivation}
\end{figure}

\section{Introduction}
\label{sec:intro}
Evaluation is shifting from static knowledge tests toward economically valuable professional work,
graded by the standards practitioners apply to one another.~\cite{miserendino2025swe,patwardhan2025gdpval,vidgen2025ai,wang2025profbench,
mazeika2025remote,akyurek2025prbench}. Financial research is a natural target for this agenda.
Analyst work informs investment decisions, diligence, reporting, governance, and
regulatory review; an incorrect source, missed adjustment, or unsupported
assumption can materially alter conclusions.
Yet financial research is not merely a final-answer task. Consider an analyst asked whether
management-adjusted EBIT is overstated if capitalized software development is treated as a
real cost. A reviewer must be able to verify the filing, fiscal period, GAAP-to-non-GAAP bridge,
treatment of amortization, and adjustment formula behind the reported figure. In practice,
the value of the answer depends on whether another analyst can audit that derivation.
Doing that work requires an analyst to identify and disambiguate entities, find primary
sources, extract line items, normalize conventions, apply accounting adjustments, and compose
a defensible number.
Recent work has separately developed rubric-based grading for open-ended answers and training
signals~\cite{deshpande2025multichallenge,arora2025healthbench,gunjal2025rubrics,
huang2025reinforcement}, making it possible to evaluate such work beyond the level of a single
final answer.

Existing finance benchmarks provide many of the domain ingredients for this agenda: numerical
reasoning over a pre-selected page~\cite{chen2021finqa,zhu2021tat,reddy2024docfinqa,
zhao2024financemath}, retrieval ranking over filings~\cite{choi2025finagentbench,
choi2025finder}, single-fact lookup against documents or structured
databases~\cite{islam2023financebench,kim2026finretrieval,wang2024redefining}, holistic
NLP-task aggregation~\cite{xie2024finben}, long-form regulatory
explanation~\cite{chen2024fintextqa}, and end-to-end agentic evaluation against expert-authored
questions~\cite{bigeard2025finance}. These abstractions are valuable because they isolate
retrieval, reading, calculation, and domain knowledge. However, they leave two parts of the
work under-specified. First, practicing analysts often receive open-ended,
multi-source, assumption-dependent questions unlike typical benchmark tasks that pair a query with
a single discrete input, such as one page, filing, retrieved chunk, or database row. Second, the work is
reviewed as a derivation, not only as a final number: source choices, time periods, line items, adjustments,
formulas, and conclusions must be verifiable by another practitioner. A benchmark for economically
valuable finance work therefore needs both practitioner-inspired question shapes and grading units
that correspond to the auditable steps of the derivation.

These two mismatches motivate a single research question:

\begin{quote}
\itshape
Can financial-research agents be evaluated on practitioner-inspired question shapes and at the
derivation-level resolution used in expert review?
\end{quote}

We introduce \textsc{BigFinanceBench} as our answer: a benchmark of $928$ full-workflow
financial-research questions written by practicing financial analysts and former
investment-banking practitioners, each audited by a separate reviewer before inclusion. The
questions reflect the open-ended, multi-source, assumption-dependent tasks these annotators
perform in practice, requiring systems to chain retrieval, normalization, accounting judgment,
and calculation rather than solve a closed-form problem over pre-selected context. We use
\emph{workflow-grounded} to mean that tasks preserve this open-ended, multi-source,
assumption-dependent structure of analyst work and are graded at the resolution used in expert
review. \textsc{BigFinanceBench} is intended as a frontier evaluation over realistic workflow
families, not as a frequency-weighted census of analyst workloads. To grade these questions at
the resolution of analyst review, each item pairs a ground-truth reference answer with a
point-weighted rubric derived from the analyst workflow. Rubric criteria decompose the
derivation into independently verifiable steps, including entity identification, source
selection, line-item retrieval, accounting adjustment, formula construction, and final synthesis.
At evaluation time, judges apply these criteria to the visible agent trajectory, including tool
calls, tool outputs, calculations, and the submitted answer, so the benchmark grades auditable
workflow evidence rather than only the final response.

\paragraph{Contributions.} We make three contributions:
\begin{itemize}
\item \textbf{Auditable finance workflows.} We introduce $928$ expert-authored
financial-research questions designed to require source selection, accounting judgment, and
calculation over public evidence.
\item \textbf{Derivation-level evaluation.} Each item includes a point-weighted rubric over
independently checkable workflow steps, yielding $15{,}656$ criteria and $36{,}241$ total points
for partial-credit evaluation of analyst derivations.
\item \textbf{Empirical diagnosis of current agents.} Across ten frontier and open-weight agents,
we show that rubric scores expose partial progress hidden by final-answer accuracy, that much of
the remaining between-model separation appears before a clean setup is reached, and that model
strengths vary across workflows.
\end{itemize}

\section{Related Work}
\label{sec:related}

\paragraph{Previous finance datasets.} Finance benchmarks have moved from curated financial
excerpts toward richer source and tool settings. Early QA datasets evaluate table-and-text
arithmetic over earnings reports and filings, with answers expressed as numbers, programs, or
normalized spans~\cite{chen2021finqa,zhu2021tat,reddy2024docfinqa}; FinanceMath adds explicit
financial concepts and formula use~\cite{zhao2024financemath}. Filing-centered benchmarks
study open-book QA, evidence retrieval, filing-type selection, and passage ranking over public
company disclosures~\cite{islam2023financebench,choi2025finder,choi2025finagentbench}.
Structured-data settings evaluate Text2API generation and agent access to fundamentals
databases~\cite{wang2024redefining,kim2026finretrieval}. Broader finance suites expand the
surface area to multi-task financial NLP, long-form regulatory QA, multilingual and multimodal
finance reasoning, filings-plus-market-data reasoning, and ReAct-style financial research
agents~\cite{xie2024finben,chen2024fintextqa,xue2024famma,agrawal2026fintradebench,
bigeard2025finance}.

\paragraph{Economically valuable and professional task evaluation.} Outside finance, evaluation
has similarly moved from knowledge tests toward work products and professional judging. MMLU
and GPQA~\cite{hendrycks2020measuring,rein2023gpqa} evaluate broad academic knowledge and
graduate-level expert science questions with multiple-choice accuracy. Economically grounded
benchmarks evaluate paid software-engineering tasks and proposal selection~\cite{miserendino2025swe},
deliverables from occupations across major GDP sectors~\cite{patwardhan2025gdpval}, and
end-to-end remote freelance projects~\cite{mazeika2025remote}. Rubric-based professional
benchmarks evaluate expert-authored prompts with source documents and criterion-level grading:
APEX covers investment banking, consulting, law, and medicine; ProfBench covers scientific and
business report generation; and PRBench covers high-stakes finance and legal
reasoning~\cite{vidgen2025ai,wang2025profbench,akyurek2025prbench}.
Table~\ref{tab:related-positioning} summarizes the closest comparison set.

\begin{table}[t]
  \centering
  \small
  \setlength{\tabcolsep}{4pt}
  \renewcommand{\arraystretch}{1.08}
  \caption{At-a-glance positioning of the closest benchmark families. Prior work provides
  finance-specific tasks, agent harnesses, or rubric-scored professional outputs; \textsc{BigFinanceBench}
  combines all three with derivation-level grading of auditable analyst workflows.}
  \label{tab:related-positioning}
  \begin{tabularx}{\linewidth}{Xcccc}
    \toprule
    Benchmark &
    \shortstack{Finance\\tasks} &
    \shortstack{Agent\\harness} &
    \shortstack{Rubric-\\scored} &
    \shortstack{Derivation-\\level} \\
    \midrule
    FinanceBench~\cite{islam2023financebench} & \cmark & \xmark & \xmark & \xmark \\
    FinRetrieval~\cite{kim2026finretrieval} & \cmark & \cmark & \xmark & \xmark \\
    Finance Agent Benchmark~\cite{bigeard2025finance} & \cmark & \cmark & \cmark & \xmark \\
    ProfBench~\cite{wang2025profbench} & \xmark & \xmark & \cmark & \xmark \\
    PRBench~\cite{akyurek2025prbench} & \cmark & \xmark & \cmark & \xmark \\
    \textsc{BigFinanceBench} & \cmark & \cmark & \cmark & \cmark \\
    \bottomrule
  \end{tabularx}
\end{table}

\paragraph{Rubric scoring and harnesses.} Recent benchmarks vary along two methodological axes:
how the model is run, and how its work is scored. Rubric-scored evaluations decompose
open-ended work into criterion-level judgments: MultiChallenge uses binary instance-level
questions, HealthBench uses physician-written criteria with positive and negative point values,
and APEX, ProfBench, and PRBench use expert-authored criteria for professional
tasks~\cite{deshpande2025multichallenge,arora2025healthbench,vidgen2025ai,wang2025profbench,
akyurek2025prbench}. Separately, agent harnesses run the model in a loop: the model observes
state, chooses whether to call a tool, receives the tool output, and repeats until it submits a
final answer or deliverable. SWE-Lancer, GDPval, RLI, BrowseComp, FinRetrieval, and Finance
Agent Benchmark instantiate this pattern with shells, editors, browsers, computer-use
interfaces, web search, SEC search, or structured-data APIs~\cite{miserendino2025swe,
patwardhan2025gdpval,mazeika2025remote,wei2025browsecomp,kim2026finretrieval,
bigeard2025finance}. Recent tool-use benchmarks also move beyond final answers by evaluating
agent trajectories, tool calls, and intermediate process quality~\cite{
lu2025toolsandbox,nath2025toolcomp,lu2025agentrewardbench}. These are close methodological
precedents, but they target generic agent behavior or judge training rather than expert
financial-research workflows. Related training work converts rubrics
into reward signals, using
prompt-specific criteria for on-policy reinforcement learning and reward-shaping~\cite{
gunjal2025rubrics,huang2025reinforcement}.

\section{The \textsc{BigFinanceBench} Dataset}
\label{sec:dataset}

\begin{figure}[h]
  \centering
  \includegraphics[width=\linewidth]{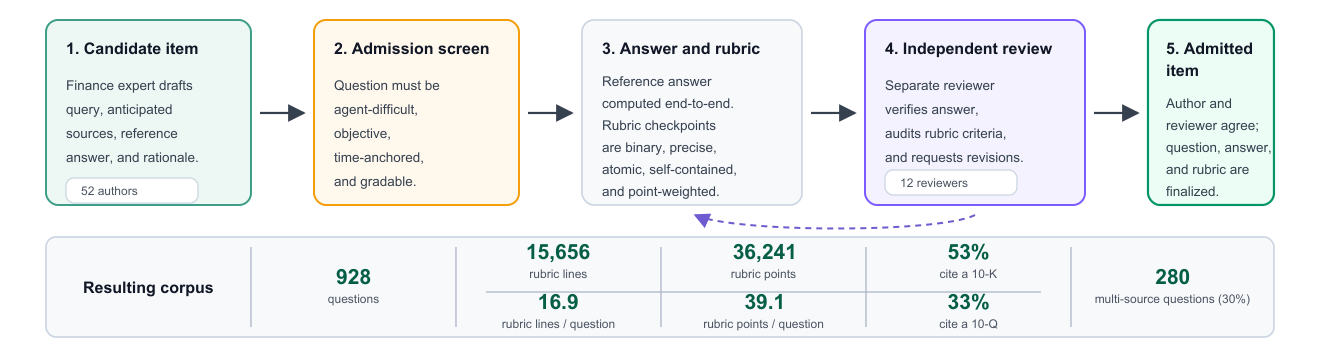}
  \caption{Dataset construction workflow and resulting-corpus statistics for
  \textsc{BigFinanceBench}.}
  \label{fig:dataset-construction}
\end{figure}

\subsection{Dataset construction}
\label{sec:dataset-construction}

The dataset contains
$928$ items authored by $52$ financial-research
subject-matter experts, predominantly current and former investment bankers and private equity
investors, with smaller representation from equity research and adjacent buy-side roles. A
separate panel of $12$ reviewers audited the items before inclusion. Each row records the
query, anticipated sources, reference answer, and weighted rubric.
The dataset was assembled between September 2025 and March 2026.

\textbf{Challenging.} Candidate questions were crafted to be purposefully difficult for existing agents. Before
submission, authors tested each query against multiple frontier agents and systems then recorded the observed failure mode(s). Reviewers accepted
questions whose difficulty came from depth of reasoning, including niche expertise,
multi-step methodology, or domain-specific judgment. They rejected questions that merely
stacked many simple lookups.

\textbf{Objective.} Candidate questions had to be objectively gradable: the answer had to be
fixed by expert consensus under a legitimate methodology, anchored to a specific point in time,
and expressed with explicit units and rounding rules. Non-obvious assumptions were stated in
the query, and rubric criteria were limited to steps reviewers deemed necessary for a
perfect answer, not artifacts of a particular solution path.

\textbf{Authoring.} For each item, the author produced a ground-truth reference answer and a point-weighted
rubric. Reference answers followed the methodology a practicing analyst would use, with
intermediate values carried through the calculation and conventional rounding applied only to
the final figure. Rubrics were governed by five authoring rules:
\begin{itemize}
  \item \textbf{Binary.} Each criterion is a yes/no determination, not a nuanced evaluation.
  \item \textbf{Precise.} Each criterion specifies the exact value and time period; the
  document is not separately tagged unless citing a specific source is itself part of the
  correct answer.
  \item \textbf{Atomic.} Each criterion isolates a single concept so that partial credit can be
  awarded fairly.
  \item \textbf{Self-contained.} Given only the candidate answer and a single rubric criterion, the
  grader can grade the criterion without reading any other criterion.
  \item \textbf{Weighted.} Each criterion carries an integer weight on a $1$ to $10$ scale, with
  the final answer always carrying a meaningful share of total points.
\end{itemize}

\textbf{Review.} Each item was routed to a separate finance expert, who independently verified
the reference answer and audited the rubric for the above five authoring rules and rounding tolerances. Authors revised items until both
parties agreed, after which the item was admitted to the training pool. In the preserved
review logs, $81\%$ of items ($748$ of $928$) contain substantive feedback or
reviewer-applied edits, consistent with meaningful audit rather than perfunctory sign-off.

Figure~\ref{fig:dataset-construction} summarizes the construction workflow and resulting corpus,
Table~\ref{tab:example-items} shows representative items, and
Appendices~\ref{app:dataset-stats} and~\ref{app:example-agent-trajectory} provide summary
statistics and a full annotated trajectory; answers are intentionally brief, while the
difficulty lies in producing the auditable derivation captured by the point-weighted rubric.
Release details are summarized in Appendix~\ref{app:limitations-release}.

\begin{table}[t]
  \caption{Example \textsc{BigFinanceBench} items with reference answers and rubric excerpts. Rubrics are clipped
  if necessary with full rubrics reported in Appendix Tables~\ref{tab:dayforce-full-rubric}
  and~\ref{tab:spotify-full-rubric}.}
  \label{tab:example-items}
  \centering
  \scriptsize
  \setlength{\tabcolsep}{2.5pt}
  \renewcommand{\arraystretch}{1.08}
  \begin{tabularx}{\linewidth}{
    >{\raggedright\arraybackslash}p{0.25\linewidth}
    >{\raggedright\arraybackslash}p{0.15\linewidth}
    >{\raggedright\arraybackslash}X}
    \toprule
    Question & Reference answer & Rubric excerpt \\
    \midrule
    If I take Dayforce's management adjusted reported EBIT as is, would it be overstated or
    understated or the same last year if I think capitalized software expense is a real cost?
    If so, by how much?
    &
    Overstated by \$90.1m of excluded capitalized software development costs. Adj EBIT was
    unburdened by any amortization of capitalized software. Subtracting capitalized software
    development spend, the adjusted management figure of \$410.5m becomes \$320.4m.
    &
    \begin{minipage}[t]{\linewidth}
      Full rubric: 30 lines, 91 points; first 10 lines shown.\par
      \texttt{[+1]} Identifies DAY as ticker\par
      \texttt{[+2]} Identifies Fiscal Year Ended December 31 2024 as the latest year\par
      \texttt{[+4]} Identifies  \$410.5m as mgmt's reported Adj. EBIT for FY24\par
      \texttt{[+2]} Identifies \$104.1m as GAAP EBIT for FY24\par
      \texttt{[+1]} Identifies stock-based compensation of \$156.6m as a component of the difference between GAAP and non-GAAP EBIT\par
      \texttt{[+2]} Identifies amortization of \$120.0m as a component of the difference between GAAP and non-GAAP EBIT\par
      \texttt{[+1]} Identifies "Other" of \$29.8m as a component of the difference between GAAP and non-GAAP EBIT\par
      \texttt{[+2]} Identifies that stock-based compensation does not include any amortization of capitalized software added back\par
      \texttt{[+1]} Identifies that "Other" includes \$19.8m of restructuring expenses\par
      \texttt{[+1]} Identifies that "Other" includes \$9.0m of fair value adjustments\par
      \textit{Clipped: remaining 20 rubric criteria omitted; full entry in Appendix
      Table~\ref{tab:dayforce-full-rubric}.}
    \end{minipage} \\
    \addlinespace
    Udemy (UDMY) Business NDRR vs. ARR Growth: Calculate the implied YoY \% Q3'26 ARR and ARR
    increase for Udemy Business using Q3 2025 financials, assuming the aggregate Net Dollar
    Retention Rate (NDRR) converges to the current 'Large Customer' benchmark and New Logo ARR
    contributes a constant \$25M. Round the final answer to two decimal places
    &
    1.74\%
    &
    \begin{minipage}[t]{\linewidth}
      Full rubric: 9 lines, 32 points; all lines shown.\par
      \texttt{[+1]} Identifies Q3 '25 Actual ARR as \$527.20M\par
      \texttt{[+3]} Identifies "Large Customer" NDRR as 97\%\par
      \texttt{[+2]} Calculates Q3 '26 Retained ARR as (Q3 '25 ARR) * (Large Customer NDRR)\par
      \texttt{[+5]} Calculates Q3 '26 Retained ARR as \$511.38M\par
      \texttt{[+1]} Identifies new logo growth as a \$25M\par
      \texttt{[+2]} CalculatesQ3 '26 ARR as Q3 '26 Retained ARR + new logo growth\par
      \texttt{[+5]} Calculates Q3 '26 ARR as \$536.38M\par
      \texttt{[+3]} Calculates ARR Growth as \$9.18M\par
      \texttt{[+10]} Calculates YoY \% ARR Increase as 1.74\%, when rounded to two decimal places
    \end{minipage} \\
    \addlinespace
    What would be the payouts to Spotify's executive officers if Spotify received a buyout
    offer at a \$800 per share price? Please use the latest disclosed incentive program
    payments (including options and RSUs), shareholdings, and change of control severance
    payments in addition to any other compensation schemes as of 9/30/2025. Please round the
    final answer to the nearest tenth of a billion.
    &
    \$23.7 billion
    &
    \begin{minipage}[t]{\linewidth}
      Full rubric: 15 lines, 31 points; first 10 lines shown.\par
      \texttt{[+5]} Calculates the payout to Spotify's executive officers by summing cash severance plus the value of all equity holdings (including options and RSUs) at the buyout price\par
      \texttt{[+3]} Utilizes the treasury stock method to calculate the net cash payment to key executives following cash required to exercise outstanding options\par
      \texttt{[+1]} Identifies Daniel Ek as a key executive officer.\par
      \texttt{[+1]} Identifies Gustave Soderstrom as a key executive officer.\par
      \texttt{[+1]} Identifies Alex Nordstrom as a key executive officer.\par
      \texttt{[+1]} Identifies Dustee Jenkins as a key executive officer.\par
      \texttt{[+1]} Identifies Christian Luiga as a key executive officer.\par
      \texttt{[+2]} Identifies a cash severance of 12x Annual Base Salary for Alex Nostrom, Cristian Luiga, and Gustav Soderstrom.\par
      \texttt{[+2]} Identifies Daniel Ek's 21.9 million outstanding share ownership (acceptable within +/- 1\%)\par
      \texttt{[+2]} Multiplies Daniel Ek's outstanding share ownership by the buyout share price to calculate payments to Daniel Ek.\par
      \textit{Clipped: remaining 5 rubric criteria omitted; full entry in Appendix
      Table~\ref{tab:spotify-full-rubric}.}
    \end{minipage} \\
    \bottomrule
  \end{tabularx}
  \renewcommand{\arraystretch}{1}
\end{table}

\begin{figure}[t]
  \centering
  \includegraphics[width=\linewidth]{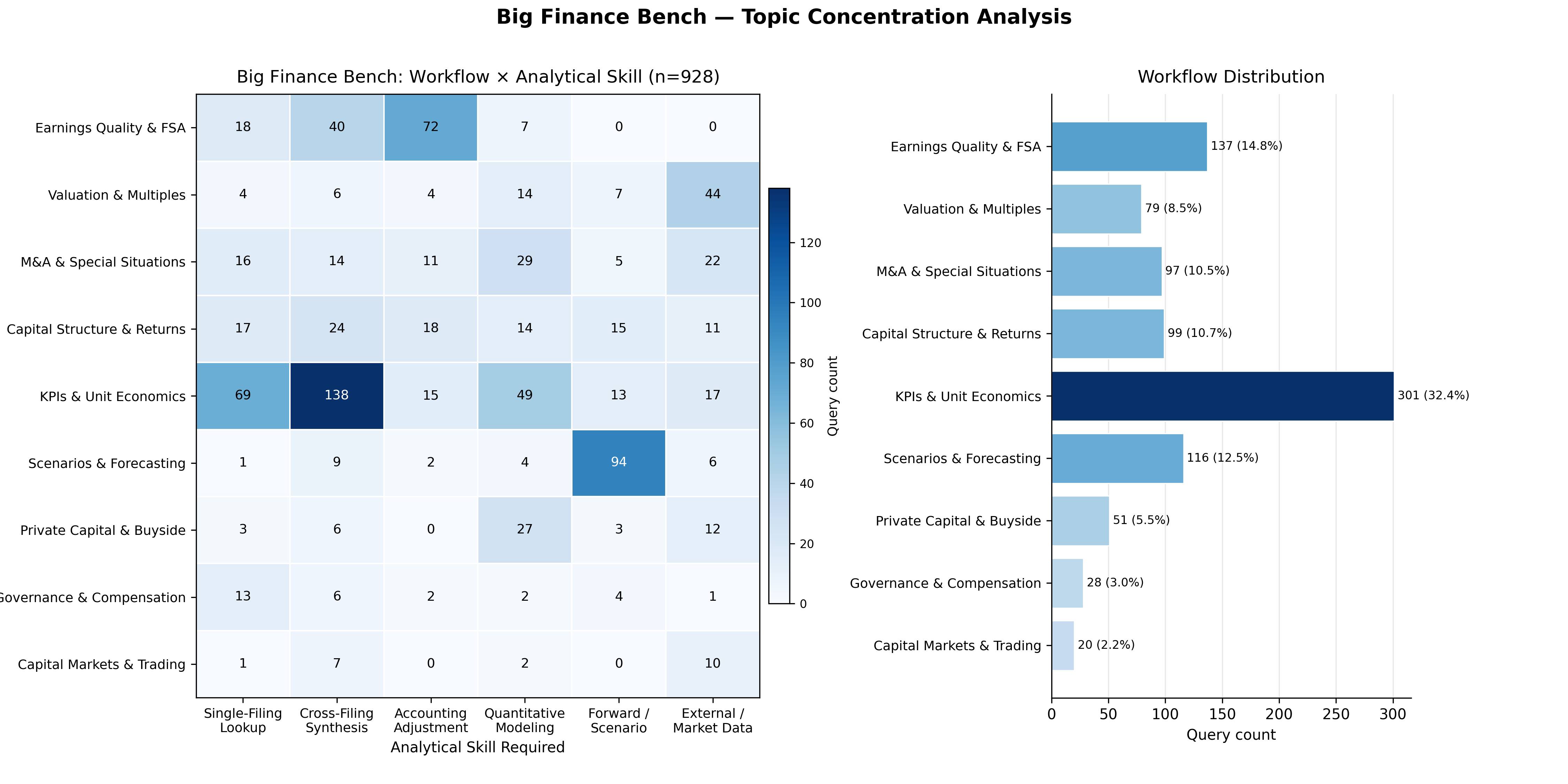}
  \caption{Workflow composition of \textsc{BigFinanceBench}.
  (a)~Joint distribution of analyst workflow type against the analytical skill
  each question requires.
  (b)~Marginal distribution over workflow types.
   See Appendix~\ref{app:workflow-classification} for additional sector coverage and question difficulty insights.}
  \label{fig:dataset-workflow}
\end{figure}

\textbf{Dataset composition.} Figure~\ref{fig:dataset-workflow} shows that \textsc{BigFinanceBench}
is centered on core public-equity research workflows: earnings quality, M\&A, valuation, and
operating KPIs. The heatmap indicates that these workflows do not collapse to a single
retrieval pattern, since short-answer items still require cross-filing synthesis,
forward/scenario reasoning, accounting adjustment, and external market data. The appendix
coverage figure (Figure~\ref{fig:dataset-sector}) shows broad but uneven sector coverage, with
heavier representation of TMT and diversified questions.
Different workflows demand different mixes of analyst work.
We categorize each rubric criterion into one of three analyst stages
(\emph{Retrieval}, \emph{Definition}, \emph{Calculation}) using the regex rules detailed in
Appendix~\ref{app:rubric-stage-classification}.
By rubric points, the stage mix is $29.5\%$ Retrieval, $5.8\%$ Definition, $63.2\%$
Calculation, and $1.5\%$ Other. Appendix Figure~\ref{fig:rubric-stage-by-workflow}
reports the corresponding per-workflow stage mix, and Appendix Figure~\ref{fig:dataset-sector}
reports sector coverage.

\section{Benchmark Evaluation}
\label{sec:benchmark-results}

\subsection{Setup}
\label{sec:benchmark-setup}

\paragraph{Agent harness.} We evaluate current frontier and open-weight models in a common
open-book ReAct-style harness with a $50$-step budget. At each step, the model may call one or
more tools or submit a final answer.
All model inference uses commercial hosted APIs; open-source models are accessed through
Vercel Model Gateway. No on-site compute workers are used for benchmark evaluation.

\paragraph{Tool surface.} The tool set is intentionally minimal and public-source only.
\texttt{web\_search} supports broad discovery, \texttt{edgar\_search} targets SEC filings,
\texttt{fetch\_url} retrieves the selected source, and \texttt{python\_exec} provides a
sandbox for calculation and tabular manipulation; \texttt{final\_answer} terminates the run.
This design tests whether agents can identify and use the relevant public evidence rather than
whether they can query a proprietary financial database.

\paragraph{Judging and metrics.} For each completed trajectory, two independent judges
(Gemini~3.1~Pro Preview and Claude Opus~4.7) receive the question, visible agent trace,
reference answer, and full point-weighted rubric, and then grade correctness for each
rubric criterion and the final answer. The primary metric is \emph{rubric score}: earned rubric points divided by total
rubric points, macro-averaged across questions. We also report \emph{final-answer accuracy}, the
fraction of final answers judged correct, and \emph{rubric criterion pass rate} (the unweighted fraction passed).
Each model is run for three trials per question. Reported metrics are computed on the paired set
of $(\text{question}, \text{trial})$ records that received a valid grade from both judges:
they average the two judges within each (question, trial), average trials within question, and
then macro-average over questions. Per-model paired coverage ranges from $2{,}776$ to
$2{,}783$ of the $2{,}784$ expected, or $99.7\%$ to $100\%$. Inter-judge agreement is high (Cohen's
$\kappa \in [0.952, 0.973]$ on final-answer correctness across the ten models); see
Appendix~\ref{app:inter-judge-kappa}.
Additional rollout and grading diagnostics are reported in
Appendix~\ref{app:harness-diagnostics}.

\subsection{Results}
\label{sec:benchmark-results-summary}

\begin{figure}[t]
  \centering
  \includegraphics[width=\linewidth]{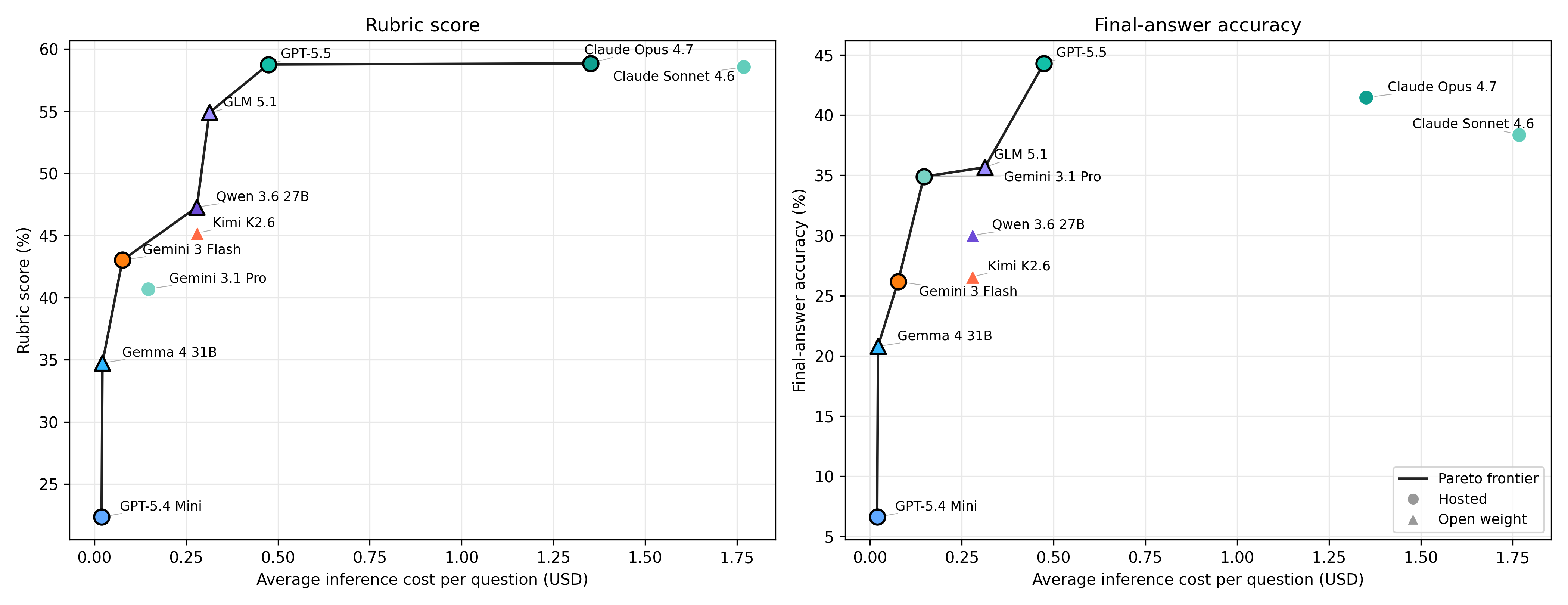}
  \caption{Performance versus cost frontier on \textsc{BigFinanceBench}. Each point plots a model's
  score against average inference cost per question, computed from trace token counts and
  provider list prices. Lines mark the metric-specific Pareto frontiers.}
  \label{fig:perf-price-pareto}
\end{figure}

\begin{figure}[t]
  \centering
  \includegraphics[width=\linewidth]{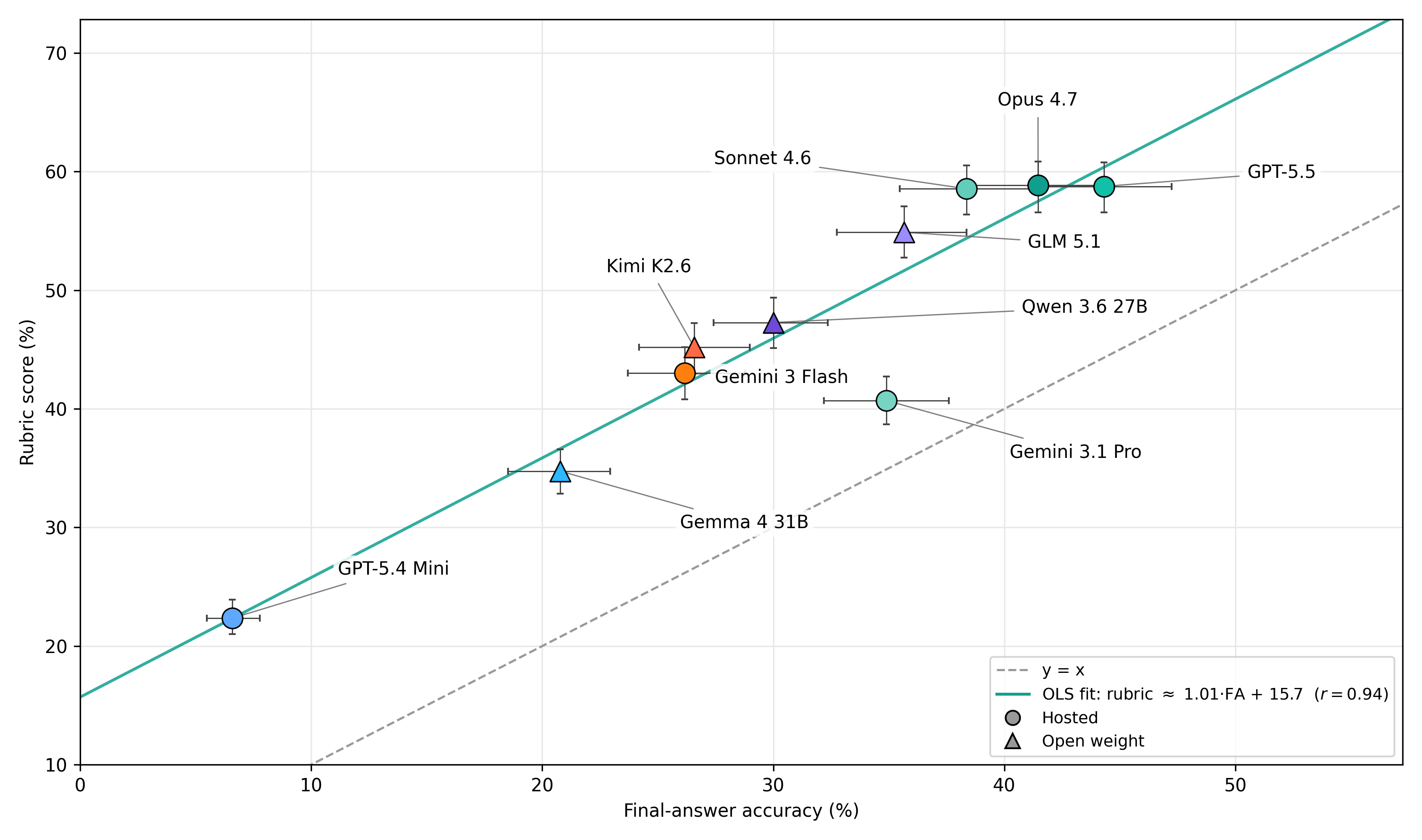}
  \caption{Per-model rubric score against final-answer accuracy. Every system sits above
  the $y = x$ diagonal (dashed grey), indicating that rubric grading credits intermediate
  workflow progress that final-answer grading discards. The teal line summarizes the overall
  trend; Gemini~3.1~Pro is the most visible deviation. Error bars are $95\%$ bootstrap
  confidence intervals over questions.}
  \label{fig:process-vs-outcome}
\end{figure}

\textsc{BigFinanceBench} remains far from saturated: the headline results figure
(Figure~\ref{fig:motivation}) shows that the best systems remain below $60\%$ rubric score and
below $45\%$ final-answer accuracy. The top models are tightly clustered on rubric despite
clearer separation in final-answer accuracy (Figure~\ref{fig:perf-price-pareto}; Appendix
Table~\ref{tab:benchmark-results}). We observe the same hardness at the question level:
rubric scoring preserves a broad middle band of partially solved questions, whereas
final-answer scoring collapses many of these items into a larger hard tail. Appendix~\ref{app:conditional-drilldowns}
reports the corresponding drill-downs, including the full heatmap
(Figure~\ref{fig:difficulty-distribution}) and source-document conditioning
(Figure~\ref{fig:source-type-accuracy}). The performance-cost frontier also depends
on the metric, so optimizing for final answers and optimizing for auditable derivations need
not select the same model.

Figure~\ref{fig:process-vs-outcome} shows that rubric and final-answer scores are strongly
aligned but systematically separated. Every model sits above the equality line, with a mean
rubric-minus-final-answer gap of $15.95$~pp, Pearson $r=0.94$ across models, and $41$ of $45$
pairwise rankings concordant. Rubric grading therefore preserves most of the ordering that
final-answer accuracy produces while adding derivation-level resolution.
One case nevertheless shows the metrics are not interchangeable:
Gemini~3.1~Pro and Gemini~3~Flash invert across the two metrics: Pro is $8.7$~pp above Flash on
final-answer accuracy, but Flash is $2.3$~pp above Pro on rubric score.

\begin{figure}[t]
  \centering
  \includegraphics[width=\linewidth]{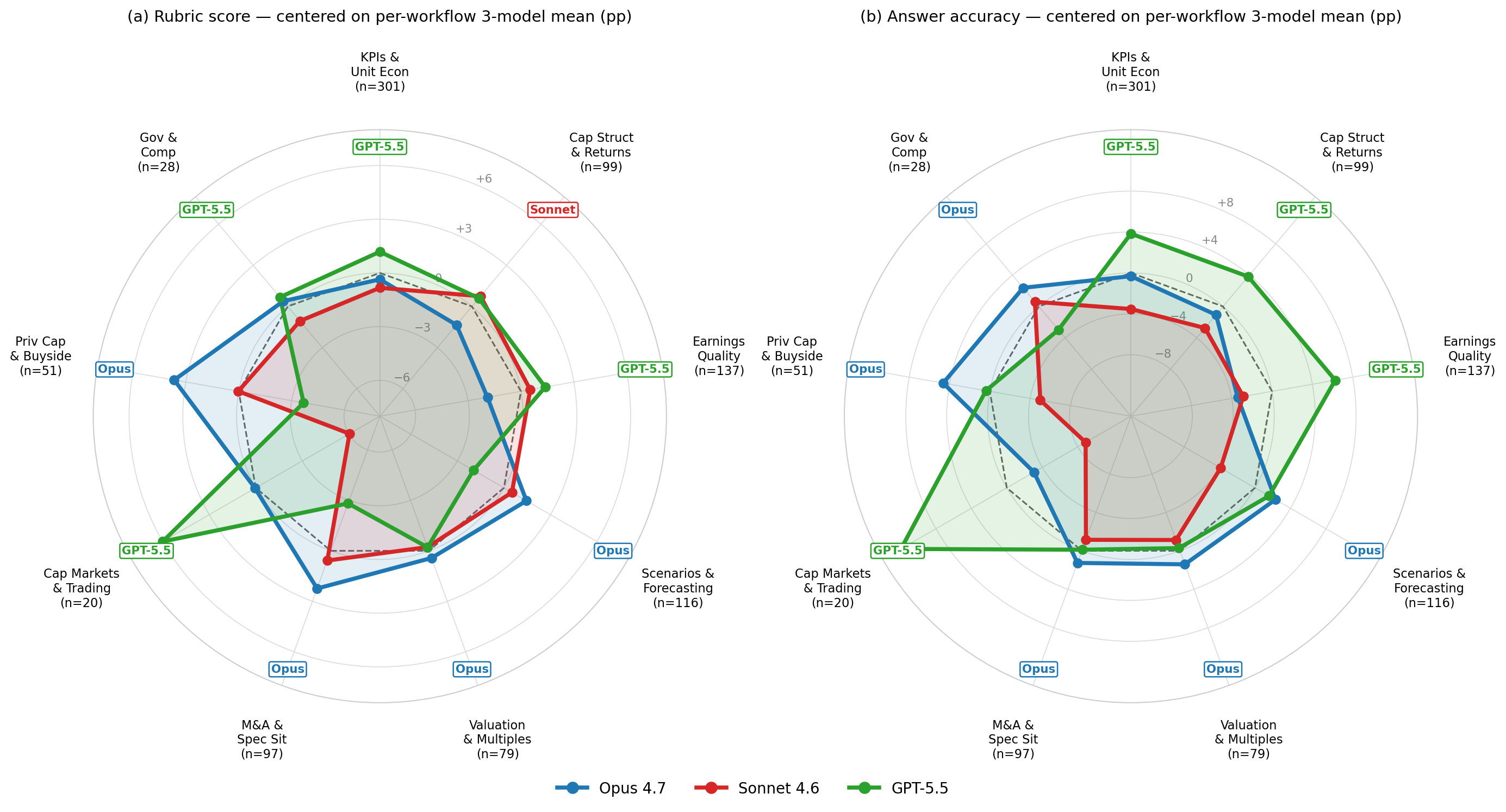}
  \caption{The headline tie among the top-3 closed models is a partition: each leads on
  different analyst workflows. Per-axis values are the model's mean rubric score (left) or
  final-answer accuracy (right) \emph{relative to the per-workflow average across the
  three models}.
  For visual clarity, we center each workflow axis by the mean of the three models, so the
  polygons highlight relative specialization rather than shared workflow difficulty.
  Polygons do not nest in either metric.
  Absolute-form polygons are
  reported in Appendix Figure~\ref{fig:partition-radar-absolute}.}
  \label{fig:partition-radar}
\end{figure}

\paragraph{Leading models specialize across workflows.}
\label{sec:partition}
The headline tie among the top-3 closed models (Opus~4.7, GPT-5.5, and Sonnet~4.6 within
$0.3$~pp of each other on rubric, Appendix Table~\ref{tab:benchmark-results}) masks different
workflow strengths. Among these three models, no model earns the highest rubric on more than
$37.7\%$ of questions under fractional tie credit ($190$ of $928$ questions are top-3 ties); each
leads on roughly a third of the benchmark. Figure~\ref{fig:partition-radar}
shows the same partition by analyst workflow: the polygons do not nest, indicating
orthogonal strengths rather than a single shared frontier.

This non-uniform frontier is also useful operationally: a simple router using observable
(workflow, source) features improves over the best single model by $7.6\%$ relative rubric
gain and $10.7\%$ relative answer-accuracy gain, while a best-of-10 oracle leaves additional
headroom (Appendix~\ref{app:routing-ladder}).

\begin{figure}[t]
  \centering
  \begin{minipage}[t]{0.49\linewidth}
    \centering
    \textbf{(a) Stage accuracy}\par\vspace{0.2em}
    \includegraphics[width=\linewidth]{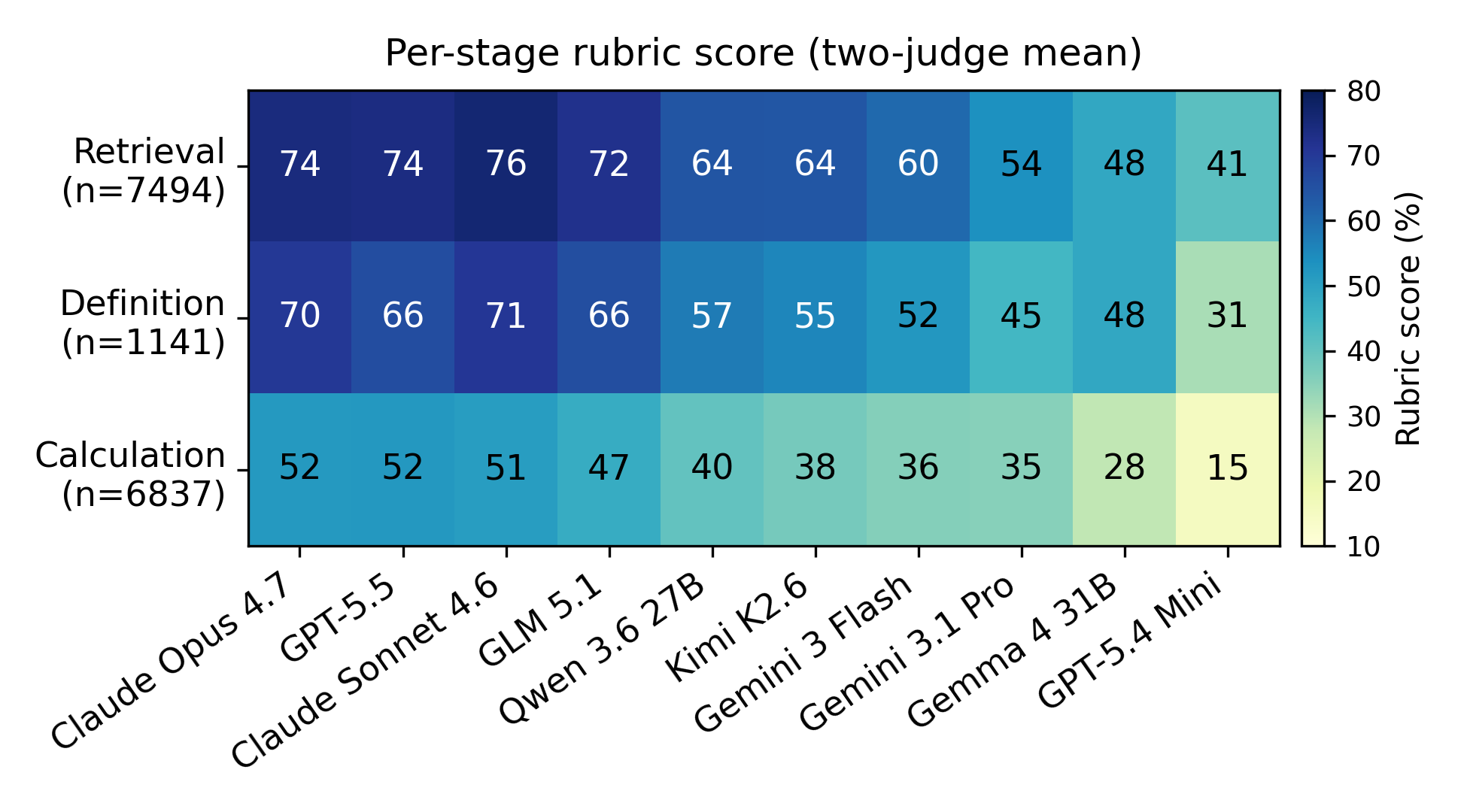}
  \end{minipage}\hfill
  \begin{minipage}[t]{0.49\linewidth}
    \centering
    \textbf{(b) Clean setup and calculation}\par\vspace{0.2em}
    \includegraphics[width=\linewidth]{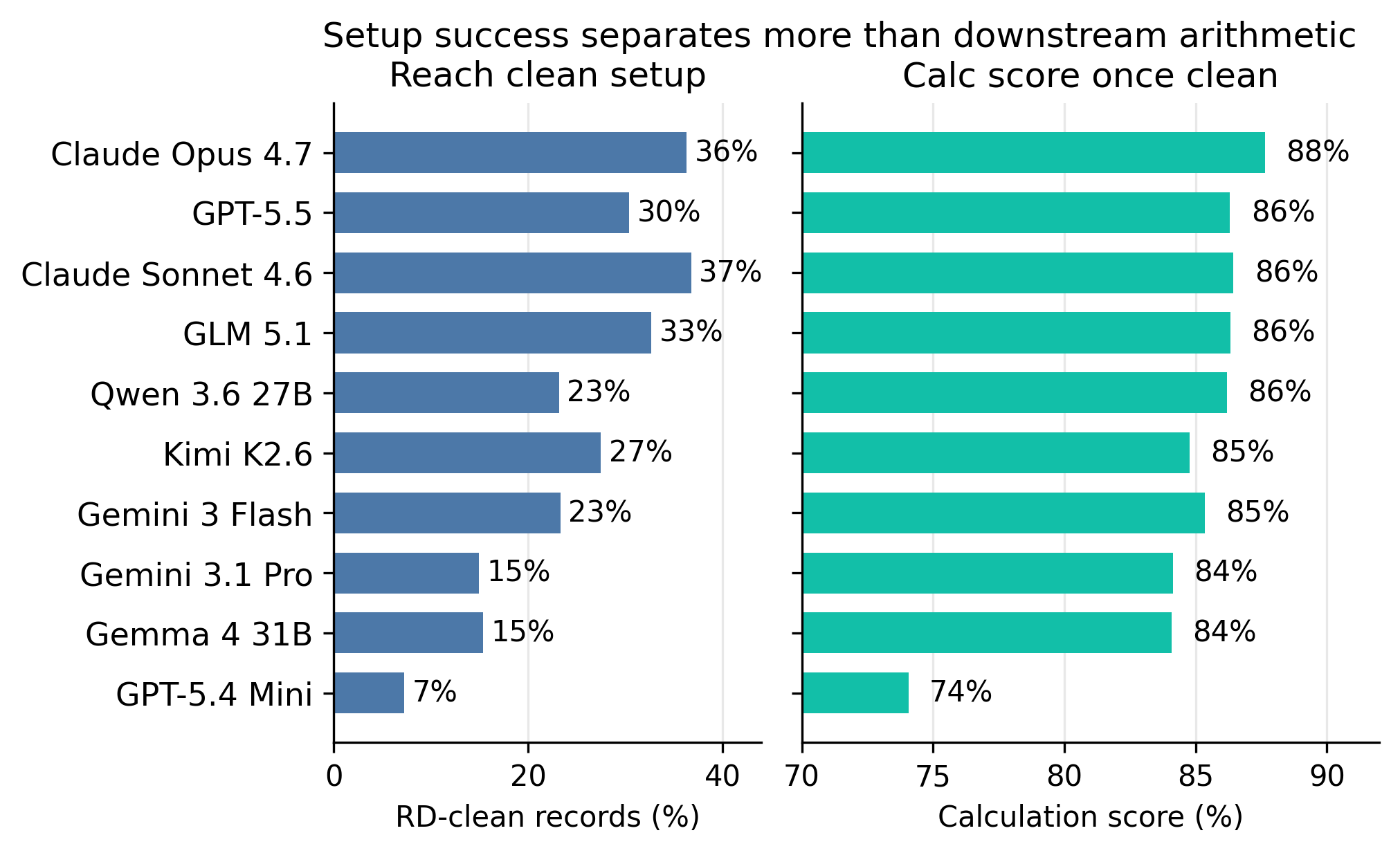}
  \end{minipage}
  \caption{Workflow-stage accuracy and clean-setup calculation. Panel (a) maps each rubric
  line to a workflow stage by its leading verb and reports the question-level macro-mean within
  questions containing that stage. Panel (b) reports the share of eligible
  $(\text{question}, \text{trial})$ records with perfect Retrieval and Definition credit, and
  the point-weighted Calculation score within that same clean-setup pool.}
  \label{fig:rubric-stage-accuracy}
  \label{fig:rd-clean-calc-compact}
\end{figure}

\paragraph{Calculation score on clean setup.}
The naive read from Panel (a) of Figure~\ref{fig:rubric-stage-accuracy}
suggests that models are weakest on calculation.
Surprisingly, the opposite is the case:
Panel (b) separates two margins. The first is whether a model reaches a clean setup: perfect
Retrieval and Definition credit. This entry rate
remains strongly model-separated, so stronger systems still complete more of the auditable
workflow.
The second margin is the point-weighted Calculation score conditional on reaching that setup,
where scores compress for most systems.
This compression is not just a denominator artifact:
Appendix~\ref{app:rd-clean-question-fe} absorbs question identity and finds much larger
question-level variation after clean setup, with a p10 to p90 span of $49.7$ pp, than residual
between-model variation ($5.1$ pp). Within this stage decomposition, most remaining
between-model separation appears before a clean setup is reached; among trajectories that
already receive perfect Retrieval and Definition credit, residual Calculation differences are
much smaller than question-to-question variation. This conditional analysis is descriptive
rather than causal, but it suggests that many failures accumulate before the final arithmetic
stage rather than arising from arithmetic alone.

\section{Conclusion}

\textsc{BigFinanceBench} makes a benchmark-methodology claim: finance agents should
be evaluated on auditable analyst workflows, not only on isolated QA, retrieval,
or arithmetic.
This follows the shift toward economically meaningful
professional work and rubric-based grading~\citep{miserendino2025swe,
mazeika2025remote,arora2025healthbench,wang2025profbench,akyurek2025prbench},
but instantiates it in a domain where the work product is a number plus its
derivation. The $928$ expert-authored questions, $15{,}656$ rubric criteria, and
$36{,}241$ rubric points measure whether a system can connect the entity,
source, period, definition, assumption, adjustment, and calculation that make a
financial answer reviewable.

The results show that this capability remains far from solved. As in
SWE-Lancer and RLI~\citep{miserendino2025swe,mazeika2025remote}, current models
complete parts of realistic professional tasks but fall short end-to-end: the
best systems remain below $60\%$ rubric score and below $45\%$ final-answer
accuracy, and the frontier is spiky rather than uniform across workflows.
Rubrics therefore complement, rather than replace, outcome evaluation: as in
HealthBench, ProfBench, and PRBench~\citep{arora2025healthbench,wang2025profbench,
akyurek2025prbench}, they expose partial progress, but in finance they also
localize whether a derivation broke at source selection, metric definition,
accounting adjustment, or calculation.
The main empirical pattern is that final-answer accuracy is a useful but lossy proxy for
workflow correctness, retrieval and setup dominate residual failures once arithmetic is
instrumented, and model capability is jagged across financial workflows. This non-uniform
frontier suggests that \textsc{BigFinanceBench} can also support future work on model routing:
different models lead on different workflows, but how to select the right model for a new
financial task remains open.

\noindent\textbf{Limitations.}
\textsc{BigFinanceBench} evaluates a static, public-source approximation of analyst work: it
does not cover proprietary databases, client interaction, or live clarification, and benchmark
scores should not be read as deployment validation for financial decisions. Rubric scores are
best interpreted as a high-resolution evaluation instrument rather than literal ground truth:
we mitigate but do not eliminate residual judge and rubric subjectivity through dual judging,
expert-authored rubrics, and independent item review. Additional limitations on scope, judging,
rubrics, and contamination are in
Appendix~\ref{app:limitations-release}.

\noindent\textbf{Release statement.}
We release a stratified $50$-question subset\footnote{\url{https://huggingface.co/datasets/RogoAI/big-finance-benchmark}} with $793$ rubric lines totaling $1{,}931$
rubric points, together with the grading harness\footnote{\url{https://github.com/Rogo-Technologies/big-finance-benchmark}} and the full $1{,}500$ agent traces for
academic research, and will maintain a hosted public
leaderboard\footnote{\url{https://bigfinancebench.com/}} for updated model results.
The full benchmark is withheld at submission time to reduce contamination and preserve future
evaluation utility; release details are in Appendix~\ref{app:limitations-release}.

\begin{ack}
We thank Strib Walker, Alex Argo, Jeevan Karamsetty and Rogo for their support, and the expert raters and
reviewers whose careful annotations and feedback made this benchmark possible.
\end{ack}

\bibliographystyle{plainnat}
\bibliography{references}

\appendix

\section{Limitations, Broader Impacts, and Release Details}
\label{app:limitations-release}

\subsection{Limitations}
\label{sec:limitations}
Four limitations apply. First, \textsc{BigFinanceBench} is a slice of financial research
rather than a complete model of the domain: it is concentrated in public-company, mostly
US-listed, English-language source work, which limits conclusions for private-market, non-US,
non-English, advisory, diligence, and client-interaction workflows. Second, the benchmark
standardizes real analyst work into a static tool-use setting: questions are time-anchored, the
harness exposes a fixed public-source tool surface, and agents cannot ask clarifying questions
or use proprietary financial databases. This makes systems comparable, but rankings may change
under a different interface or live analyst environment. Third, evaluation remains
approximate. We mitigate judge noise by grading each trajectory with two independent LLM
judges, reporting inter-judge agreement, and manually checking samples for systematic failure
modes such as rejection of real-world auditability requirements; nevertheless, judge scores are
not ground truth and rubric-criterion credit can differ (Appendix~\ref{app:inter-judge-kappa}).
Similarly, rubrics are reviewed by a separate finance expert and intended to contain steps
necessary for a perfect answer, but residual subjectivity remains in any finite decomposition
of an analyst-correct derivation. Fourth, the benchmark is vulnerable to contamination and
temporal drift. We time-anchor questions to reduce ambiguity from later filings, but
public-source facts can enter model training corpora over time.

\subsection{Broader Impacts}
\label{sec:broader-impacts}
\textsc{BigFinanceBench} lowers the barrier for academic and industrial groups to evaluate
financial-research agents on auditable workflows, which we view as necessary for safer and more
transparent use of such systems. However, benchmark performance should not be read as an
endorsement of financial automation or as financial, legal, or investment advice. Misuse could
support unreliable analysis, overreliance in high-stakes decisions, market-relevant
misinformation, or attempts to influence markets; deployment therefore requires
domain-specific validation, expert review, compliance review, and contextual assessment beyond
this benchmark.

\subsection{Release Details}
\label{sec:data-availability}
We release a $50$-question subset comprising questions, reference answers, rubrics, and the
grading harness for academic research. The subset is stratified by workflow and difficulty
quartile and preserves full-benchmark model rankings under rubric score (Kendall's
$\tau = 0.956$). All released annotations were author-generated and are distributed under CC
BY 4.0. The public subset\footnote{\url{https://huggingface.co/datasets/RogoAI/big-finance-benchmark}}
and evaluation harness\footnote{\url{https://github.com/Rogo-Technologies/big-finance-benchmark}}
are available for academic research.
Updated model results will be maintained on the project
website\footnote{\url{https://bigfinancebench.com/}}.
Items are grounded in publicly available financial materials and were screened for
personally identifiable information and sensitive customer data. The full benchmark is withheld
at submission time to mitigate benchmark contamination and preserve evaluation validity, but
we will consider bona fide research requests for additional access under conditions that
preserve evaluation validity.

\section{Detailed Dataset Construction}
\label{app:detailed-dataset-construction}

The current master release contains $928$ items, authored by a panel of $52$ subject-matter
experts, predominantly current and former investment bankers and private equity investors,
with smaller representation from equity research and adjacent buy-side roles, and audited by
$12$ reviewers. Reviewer involvement is non-trivial: the busiest single reviewer audited over
a quarter of the dataset. Each row records the natural-language query, the author's free-text
reasoning notes on what makes the question difficult, the anticipated source documents, the
reference answer, the weighted rubric, an optional screenshot or hyperlink to a backup source,
and the reviewer's name and review log.

Authors were asked to write items that were both \emph{challenging} and \emph{objective}, and
these two criteria were operationalized into specific rules that gated admission to the
dataset.

\textbf{Challenging.} A query qualified only if our in-house financial-research agent and at
least one other frontier system (Claude, ChatGPT, or Gemini) failed to answer it correctly when
the author tested it before submission, and the author recorded the specific failure mode in a
free-text notes field. We deliberately distinguished two failure modes and accepted only one.
The accepted mode is complexity through \emph{depth of reasoning}: queries that fail because
the model lacks the niche expertise, multi-step methodology, or domain-specific judgment to
answer correctly (e.g.\ stripping Ford Credit out of Ford's TEV/EBIT, summing non-employee
director shares from a proxy statement, or decomposing a hospital revenue bridge into available
patient days, occupancy, and the rate-volume covariance). The rejected mode is complexity
through \emph{stacked simplicity}: asking the same simple lookup repeatedly across many
entities or periods, where wrong answers come from hallucinations on trivial steps rather than
from genuine reasoning failure. Reviewers steered authors away from this pattern, and query
diversity was expected to come from varied methodology rather than from changing the company
name on a template.

\textbf{Objective.} The objectivity criterion was operationalized as three rules. (i)
\emph{Expert consensus}: the reference answer itself had to be objective, meaning any finance
expert applying a legitimate methodology would arrive at the same final figure, so open-ended
or opinion questions were excluded. For rubric criteria, authors were given a $95$-of-$100$
heuristic to apply when deciding whether a step belonged in the rubric: include a step only if
at least $95$ of $100$ finance experts would agree it belongs in any perfect answer to the
query, irrespective of which legitimate methodology the expert applied. This guidance steered
authors away from methodology-specific intermediate steps, since many calculations admit
multiple valid approaches (e.g., LTM revenue can be derived several ways) and an expert taking
a different valid approach would legitimately skip the step. The $95$-of-$100$ standard was a
heuristic for authors and reviewers to think with, not an independently tested or audited
threshold. (ii) \emph{Time-anchored}: the query had to reference a specific point in time so
the correct answer would not change months later; phrasings like ``the latest quarter'' or
``the last three fiscal years'' were disallowed. (iii) \emph{Binary verifiability}: the
response had to be gradable yes-or-no against a clear rubric, which in practice meant final
answers were currencies, multiples, percentages, or basis points with units and rounding rules
stated explicitly in the query itself. Any non-obvious assumption needed to answer the query
had to be written into the query text rather than left as something the model was expected to
infer.

\textbf{Authoring.} For each item the author produced two artifacts. The \emph{perfect answer}
is a definitive response written with the methodology a practicing analyst would actually use,
calculated end-to-end without rounding intermediate values and applying conventional rounding
only to the final figure. The \emph{rubric} is a weighted checklist of binary criteria
governed by five authoring rules:
\begin{itemize}
  \item \textbf{Binary.} Each criterion is a yes/no determination, not a nuanced evaluation.
  \item \textbf{Precise.} Each criterion specifies the exact value and time period (e.g.,
  ``share count of $1{,}622{,}843{,}689$ as of AMD's Q2'25 reporting'' rather than ``share
  count from the latest filing''); the document is not separately tagged unless citing a
  specific source is itself part of the correct answer.
  \item \textbf{Atomic.} Each criterion covers a single concept; a $5$-year CAGR, for example,
  is split into separate criteria for each input, the formula, and the final result so that
  partial credit is awarded fairly.
  \item \textbf{Self-contained.} Given only the candidate answer and a single rubric criterion, the
  grader can grade the criterion without reading any other criterion; phrasings like
  ``uses the above calculation'' or ``does the addition correctly'' were disallowed because
  they require cross-criterion context.
  \item \textbf{Weighted.} Each criterion carries an integer weight on a $1$ to $10$ scale,
  with the final answer always carrying a meaningful share of total points.
\end{itemize}
Where intermediate calculations involved repeating decimals or irrational values, authors
specified an explicit margin of error and verified that any value in the allowed range still
rounded to the correct final answer.

\textbf{Review.} Each item was routed to a separate reviewer, a different finance expert who
had not authored the item. The reviewer (i) independently performed the underlying calculation
to verify the reference answer rather than taking the author's number on trust, and (ii)
audited the rubric for objectivity, atomicity, self-containment, value precision, point
distribution, and rounding tolerances. Comments were returned to the author, who edited and
resubmitted, and the reviewer re-checked. The loop continued until both parties agreed, at
which point the item's status advanced to ``Ready for grader model'' and the row was admitted
to the training pool; all $928$ rows in the current master release reached this state.

The review log preserved on each row reflects how active this loop was: $81\%$ of items ($748$
of $928$) carry substantive feedback or reviewer-applied edits, with the remaining $19\%$
signed off with pure acknowledgment (e.g.\ ``looks good'' or ``ready to go''). This figure is a
lower bound on the true rework rate: the log preserves only the final state of the comment
thread, and authors and reviewers often condensed earlier substantive back-and-forth into a
single closing ``looks good'' line once the item was accepted. Early in the project, reviewer
feedback focused on structural rubric construction; over time, as authors internalized the
format, feedback shifted toward content, domain depth, point weighting, and technical opinions
on methodology, and review served less as quality control and more as a second-expert
sign-off.

\textbf{Author engagement and compensation.} Authors and reviewers were not anonymous
crowdworkers but finance professionals, including current and former investment bankers, private
equity investors, equity research analysts, and adjacent buy-side practitioners, all
required to have high finance backgrounds, predominantly in banking and private equity.
They were engaged through a dedicated staffing partner and through direct outreach across
the company's and existing experts' professional networks in investment banking, private
equity, and MBA programs. Contributors were predominantly US-based, with some having ties
to other countries, and were compensated at competitive professional hourly rates well
above the prevailing US minimum wage, consistent with the NeurIPS Code of Ethics.
The full set of instructions given to authors is reproduced above in this appendix, including the
admission criteria (\emph{challenging} and \emph{objective}), the five rubric authoring
rules, the review loop, and worked examples drawn from previously accepted
items; no separate participant-facing instructions or screenshots are reproduced here
because authoring took place via private contracting rather than on a public crowdsourcing
platform.

\textbf{Human subjects and risk assessment.} The data collection effort did not constitute
human subjects research in the usual sense: authors produced finance content (queries,
reference answers, rubrics) rather than serving as study subjects, and no behavioral,
psychological, biometric, or otherwise personal data was collected from them beyond the
professional credentials and authorship attribution preserved for provenance. Working
conditions were those of ordinary remote contract work, and no disclosure of medical,
financial, or other sensitive personal information was solicited. On that basis, IRB review
was determined not applicable and not sought; no risks to participants beyond those of
ordinary professional employment were identified, and none were disclosed to participants
because none were anticipated.

\section{Additional Tables and Figures}
\label{app:additional-tables-figures}

\begin{table}[t]
  \caption{Model results on \textsc{BigFinanceBench} ($n = 928$ questions $\times\,3$ trials per
  model; metrics are the mean across two independent judges, Gemini~3.1~Pro Preview and
  Claude Opus~4.7). ``Rubric score'' is the macro-averaged fraction of point-weighted rubric
  credit earned. ``Answer acc.'' is the fraction of final answers judged correct. ``Rubric
  criteria'' is the macro-averaged fraction of unweighted rubric-criterion credit earned. $95\%$
  bootstrap confidence intervals (over questions) are within $\pm 2.3$ percentage points for
  rubric score, $\pm 3.1$ for answer accuracy, and $\pm 2.2$ for rubric criteria.}
  \label{tab:benchmark-results}
  \centering
  \small
  \begin{tabular}{lrrr}
    \toprule
    Model & Rubric score & Answer acc. & Rubric criteria \\
    \midrule
    \texttt{Claude Opus 4.7} & $58.8\%$ & $41.5\%$ & $66.6\%$ \\
    \texttt{GPT-5.5} & $58.8\%$ & $44.3\%$ & $65.3\%$ \\
    \texttt{Claude Sonnet 4.6} & $58.5\%$ & $38.4\%$ & $67.3\%$ \\
    \texttt{GLM 5.1} & $54.9\%$ & $35.7\%$ & $63.2\%$ \\
    \texttt{Qwen 3.6 27B} & $47.3\%$ & $30.0\%$ & $55.3\%$ \\
    \texttt{Kimi K2.6} & $45.2\%$ & $26.6\%$ & $53.7\%$ \\
    \texttt{Gemini 3 Flash} & $43.0\%$ & $26.2\%$ & $50.7\%$ \\
    \texttt{Gemini 3.1 Pro} & $40.7\%$ & $34.9\%$ & $45.3\%$ \\
    \texttt{Gemma 4 31B} & $34.7\%$ & $20.8\%$ & $41.6\%$ \\
    \texttt{GPT-5.4 Mini} & $22.3\%$ & $6.6\%$ & $30.4\%$ \\
    \bottomrule
  \end{tabular}
\end{table}

\begin{table}[p]
  \caption{Full Dayforce example item from Table~\ref{tab:example-items}.}
  \label{tab:dayforce-full-rubric}
  \centering
  \scriptsize
  \setlength{\tabcolsep}{2.5pt}
  \renewcommand{\arraystretch}{1.08}
  \begin{tabularx}{\linewidth}{
    >{\raggedright\arraybackslash}p{0.22\linewidth}
    >{\raggedright\arraybackslash}p{0.20\linewidth}
    >{\raggedright\arraybackslash}X}
    \toprule
    Question & Reference answer & Full rubric \\
    \midrule
    If I take Dayforce's management adjusted reported EBIT as is, would it be overstated or
    understated or the same last year if I think capitalized software expense is a real cost?
    If so, by how much?
    &
    Overstated by \$90.1m of excluded capitalized software development costs. Adj EBIT
    was unburdened by any amortization of capitalized software. Subtracting capitalized
    software development spend, the adjusted management figure of \$410.5m becomes \$320.4m.
    &
    \begin{minipage}[t]{\linewidth}
      Full rubric: 30 lines, 91 points.\par
      \texttt{[+1]} Identifies DAY as ticker\par
      \texttt{[+2]} Identifies Fiscal Year Ended December 31 2024 as the latest year\par
      \texttt{[+4]} Identifies  \$410.5m as mgmt's reported Adj. EBIT for FY24\par
      \texttt{[+2]} Identifies \$104.1m as GAAP EBIT for FY24\par
      \texttt{[+1]} Identifies stock-based compensation of \$156.6m as a component of the difference between GAAP and non-GAAP EBIT\par
      \texttt{[+2]} Identifies amortization of \$120.0m as a component of the difference between GAAP and non-GAAP EBIT\par
      \texttt{[+1]} Identifies "Other" of \$29.8m as a component of the difference between GAAP and non-GAAP EBIT\par
      \texttt{[+2]} Identifies that stock-based compensation does not include any amortization of capitalized software added back\par
      \texttt{[+1]} Identifies that "Other" includes \$19.8m of restructuring expenses\par
      \texttt{[+1]} Identifies that "Other" includes \$9.0m of fair value adjustments\par
      \texttt{[+1]} Identifies that "Other" includes \$1.0m of receivables securitization fees\par
      \texttt{[+1]} Concludes that the "Other" of \$29.8m is comprised entirely of restructuring expenses, fair value adjustments, and receivables securitization fees\par
      \texttt{[+2]} Concludes that "Other" does not include any amortization of capitalized software development expense\par
      \texttt{[+4]} Notes that the amortization added back to get to non-GAAP EBIT includes some acquisition-related intangible assets\par
      \texttt{[+4]} Concludes that this amortization of intangibles must be included in the \$120.0m of amortization.\par
      \texttt{[+2]} Confludes that further searching has to be done to confirm whether the \$120.0m includes capitalized software development expense\par
      \texttt{[+4]} Identifies amortization of capitalized software development expense of \$70.7m\par
      \texttt{[+3]} Calculates 120.0 - 70.7 = 49.3m of other unexplained amortization added back\par
      \texttt{[+3]} Identifies that the \$120.0m of amortization is comprised of \$70.7m of amortization of capitalized software and 49.3m of other unexplained amortization\par
      \texttt{[+2]} Updates non-GAAP EBIT bridge to show GAAP EBIT, + stock based comp, plus amortization of capitalized software, plus other amortization, plus other, equals non-GAAP EBIT.\par
      \texttt{[+2]} Searches for other sources of amortization in the 10-K that explain the 49.3 of unexplained amortization\par
      \texttt{[+2]} Searches other SEC filings for FY24 amortization decompositions\par
      \texttt{[+2]} Notes that not enough information is provided to confidently identify the source of the remaining \$49.3m of amortization\par
      \texttt{[+1]} Notes that Ceridian Trade Name depreciation likely comprised a meaningful amount of the other amortization in FY24\par
      \texttt{[+7]} Confirms that reported non-GAAP EBIT figure is not burdened by amortization of capitalized software expense\par
      \texttt{[+7]} Identifies FY24 capitalized software expense of \$90.1m\par
      \texttt{[+10]} Notes that non-GAAP EBIT is overstated by \$90.1m of capitalized software development expense\par
      \texttt{[+7]} Subtracts capitalized software expense from non-GAAP EBIT\par
      \texttt{[+5]} Concludes that the more appropriate non-GAAP EBIT figure is \$320.4m\par
      \texttt{[+5]} Cites Dayforce 2024 10K for various figures
    \end{minipage} \\
    \bottomrule
  \end{tabularx}
  \renewcommand{\arraystretch}{1}
\end{table}

\begin{table}[p]
  \caption{Full Spotify example item from Table~\ref{tab:example-items}.}
  \label{tab:spotify-full-rubric}
  \centering
  \scriptsize
  \setlength{\tabcolsep}{2.5pt}
  \renewcommand{\arraystretch}{1.08}
  \begin{tabularx}{\linewidth}{
    >{\raggedright\arraybackslash}p{0.22\linewidth}
    >{\raggedright\arraybackslash}p{0.20\linewidth}
    >{\raggedright\arraybackslash}X}
    \toprule
    Question & Reference answer & Full rubric \\
    \midrule
    What would be the payouts to Spotify's executive officers if Spotify received a buyout
    offer at a \$800 per share price? Please use the latest disclosed incentive program
    payments (including options and RSUs), shareholdings, and change of control severance
    payments in addition to any other compensation schemes as of 9/30/2025. Please round the
    final answer to the nearest tenth of a billion.
    &
    \$23.7 billion
    &
    \begin{minipage}[t]{\linewidth}
      Full rubric: 15 lines, 31 points.\par
      \texttt{[+5]} Calculates the payout to Spotify's executive officers by summing cash severance plus the value of all equity holdings (including options and RSUs) at the buyout price\par
      \texttt{[+3]} Utilizes the treasury stock method to calculate the net cash payment to key executives following cash required to exercise outstanding options\par
      \texttt{[+1]} Identifies Daniel Ek as a key executive officer.\par
      \texttt{[+1]} Identifies Gustave Soderstrom as a key executive officer.\par
      \texttt{[+1]} Identifies Alex Nordstrom as a key executive officer.\par
      \texttt{[+1]} Identifies Dustee Jenkins as a key executive officer.\par
      \texttt{[+1]} Identifies Christian Luiga as a key executive officer.\par
      \texttt{[+2]} Identifies a cash severance of 12x Annual Base Salary for Alex Nostrom, Cristian Luiga, and Gustav Soderstrom.\par
      \texttt{[+2]} Identifies Daniel Ek's 21.9 million outstanding share ownership (acceptable within +/- 1\%)\par
      \texttt{[+2]} Multiplies Daniel Ek's outstanding share ownership by the buyout share price to calculate payments to Daniel Ek.\par
      \texttt{[+1]} Recognizes that Daniel Ek is leaving the CEO position at year-end 2025 and thus does not receive cash severence.\par
      \texttt{[+2]} Calculates the total payout for Daniel Ek of \$23.3 billion (acceptable within +/- 1\%)\par
      \texttt{[+2]} Calculates the total payout for Gustav Söderström of \$223.7 million (acceptable within +/- 1\%)\par
      \texttt{[+2]} Calculates the total payout for Alex Norström of \$148.8 million (acceptable within +/- 1\%)\par
      \texttt{[+5]} Final calculated grand total payout to the executive team is \$23.7 billion (acceptable within +/- 1\%)
    \end{minipage} \\
    \bottomrule
  \end{tabularx}
  \renewcommand{\arraystretch}{1}
\end{table}

\section{Example Agent Trajectory}
\label{app:example-agent-trajectory}

Table~\ref{tab:example-agent-trajectory} shows a complete successful trajectory for the Udemy
question in Table~\ref{tab:example-items}. The run is from Claude Opus~4.7 and illustrates a
successful EDGAR path: the agent finds the Q3 2025 $8$-K, follows the SEC filing directory to
the press-release exhibit, extracts the ARR and NDRR inputs, and computes the projection. The
harness recorded visible assistant text and tool calls, but no private chain of thought
(\texttt{thinking=off}). Long URLs and query strings are shortened for readability; every tool
call in the trajectory is kept on its own line.

\begin{table}[p]
  \caption{Complete successful Claude Opus~4.7 trajectory for the Udemy question in
  Table~\ref{tab:example-items}. The model submitted $1.74\%$ YoY growth, a $\$9.18$M ARR
  increase, and $\$536.38$M implied Q3'26 ARR; both judges marked the final answer correct and
  awarded $32/32$ rubric points ($9/9$ lines).}
  \label{tab:example-agent-trajectory}
  \centering
  \scriptsize
  \setlength{\tabcolsep}{2.5pt}
  \renewcommand{\arraystretch}{1.08}
  \begin{tabularx}{\linewidth}{
    >{\raggedright\arraybackslash}p{0.06\linewidth}
    >{\raggedright\arraybackslash}p{0.34\linewidth}
    >{\raggedright\arraybackslash}p{0.34\linewidth}
    >{\raggedright\arraybackslash}X}
    \toprule
    Step & Visible agent text and tool call & Key returned content & Reward signal \\
    \midrule
    0a &
    \emph{``I'll research Udemy's Q3 2025 financials to find the relevant metrics.''}\par
    \texttt{edgar\_search(ticker=UDMY, form\_type=8-K)}
    &
    EDGAR returns recent UDMY $8$-Ks, including the 2025-10-29 Q3 2025 results filing.
    &
    Source discovery. \\
    0b &
    \texttt{web\_search} with query: Udemy Q3 2025 earnings Udemy Business ARR NDRR large
    customer
    &
    Result snippets report Udemy Business NDRR of $93\%$ and Large Customer NDRR of $97\%$.
    &
    \cmark{} \texttt{[+3]} Large Customer NDRR is $97\%$. \\
    1 &
    \emph{``Let me get the Q3 2025 details to find the exact ARR figure.''}\par
    \texttt{fetch\_url(url=Udemy static Q3 2025 file, query=UB ARR)}
    &
    Fetch failed.
    &
    No direct credit. \\
    2 &
    \texttt{fetch\_url(url=Udemy investor-relations release, query=UB ARR and NDRR)}
    &
    Fetch failed.
    &
    No direct credit. \\
    3 &
    \texttt{fetch\_url(url=SEC Q3 2025 8-K, query=UB ARR and NDRR)}
    &
    Fetches the SEC $8$-K shell for the 2025-10-29 Q3 results filing.
    &
    Source discovery. \\
    4a &
    \emph{``I need the press release exhibit. Let me check the filing index.''}\par
    \texttt{fetch\_url(url=SEC company 8-K index, query=October 29 2025 press release)}
    &
    Returns the SEC filing index containing the Q3 2025 $8$-K accession.
    &
    Source discovery. \\
    4b &
    \texttt{web\_search} with query: Udemy Q3 2025 Udemy Business ARR
    &
    Supplemental search result points back to Udemy investor materials.
    &
    No direct credit. \\
    5 &
    \texttt{fetch\_url(url=SEC accession directory, query=press release exhibit 99.1)}
    &
    Returns the SEC directory for accession \texttt{0001607939-25-000138}.
    &
    Source discovery. \\
    6 &
    \texttt{fetch\_url(url=SEC q32025pressrelease.htm, query=UB ARR and NDRR)}
    &
    SEC exhibit reports UB ARR of \$527.2M and Large Customer NDRR of $97\%$.
    &
    \cmark{} \texttt{[+1]} Q3 '25 ARR is \$527.20M.\par
    Confirms \texttt{[+3]}. \\
    7 &
    \emph{``Now let's compute the implied Q3'26 ARR using NDRR = 97\% applied to existing ARR
    + \$25M new logo ARR.''}\par
    \texttt{python\_exec} with \texttt{q3\_25\_arr=527.2}, \texttt{ndrr=0.97}, and
    \texttt{new\_logo=25.0}
    &
    \texttt{Q3'26 ARR: \$536.38M}\par
    \texttt{ARR Increase: \$9.18M}\par
    \texttt{YoY Growth: 1.74\%}
    &
    \cmark{} \texttt{[+2]} retained ARR formula.\par
    \cmark{} \texttt{[+1]} new logo ARR is \$25M.\par
    \cmark{} \texttt{[+2]} Q3 '26 ARR formula.\par
    \cmark{} \texttt{[+5]} Q3 '26 ARR is \$536.38M.\par
    \cmark{} \texttt{[+3]} ARR growth is \$9.18M.\par
    \cmark{} \texttt{[+10]} YoY increase is $1.74\%$. \\
    8 &
    \texttt{final\_answer} with answer: Q3'26 UB ARR \$536.38M; ARR increase \$9.18M;
    YoY growth $1.74\%$.
    &
    Submitted answer includes the retained-ARR intermediate: \$527.2M $\times$ 0.97 =
    \$511.38M.
    &
    \cmark{} \texttt{[+5]} retained ARR is \$511.38M.\par
    \cmark{} Final answer correct. \\
    \bottomrule
  \end{tabularx}
  \renewcommand{\arraystretch}{1}
\end{table}

\section{Clean-Setup Calculation Fixed Effects}
\label{app:rd-clean-question-fe}

A complementary way to remove question-mix effects is to keep all RD-clean
records and absorb question identity directly. Let $R_{mqt}$, $D_{mqt}$, and
$C_{mqt}$ denote the Retrieval, Definition, and Calculation scores for model
$m$, question $q$, and trial $t$, again pooling the two judges at the
rubric-point level. We restrict to records with positive possible points in all
three stages and with $R_{mqt}=D_{mqt}=1$. We exclude GPT-5.4~Mini from this fit
because its RD-clean pool is small and its conditional Calculation score is a
lower-tail outlier; the fixed-effects model asks whether the middle-pack
compression survives question adjustment:
\[
  C_{mqt} = \alpha_q + \beta_m + \epsilon_{mqt}.
\]
The question fixed effect $\alpha_q$ gives each question its own baseline
Calculation score. The model effect $\beta_m$ is therefore estimated from
within-question contrasts: among records where models reached a clean setup on
the same question, which model received more Calculation credit? We weight each
record by its available Calculation points, matching the point-pooled
conditional score used in Figure~\ref{fig:rd-clean-calc-compact}. The
graph over models and questions is connected, so the model effects are comparable on one
scale. The fitted design uses $2{,}909$ RD-clean observations. The relevant
diagnostic is not overall fit quality but the relative scale of question and
model variation after setup is clean. The centered question effects have
standard deviation $22.1$ percentage points and a p10 to p90 span of $49.7$
percentage points, while the centered model effects range from $-2.3$ to $+2.8$
percentage points, a $5.1$ percentage-point span. The conditional calculation
pool is therefore dominated by question-level difficulty; after comparing
models within the same questions, residual model separation is much smaller.

Table~\ref{tab:rd-clean-question-fe} shows the result. Before adjustment, the
point-weighted conditional Calculation scores for these nine models span
$3.6$ percentage points ($84.1\%$ to $87.6\%$). After absorbing question
identity, the centered model effects span $5.1$ percentage points. Thus question
composition masks a small amount of residual middle-pack separation, but it does
not turn the conditional result back into the large unconditional Calculation
gap. Once setup is clean, the middle systems remain close.

\begin{table}[h]
  \caption{Question fixed-effects check for Calculation score among
  Retrieval+Definition-clean records, excluding GPT-5.4~Mini from the fit.
  ``Raw calc'' is the point-weighted conditional Calculation score. The
  fixed-effects column is the centered model effect from
  $C_{mqt}=\alpha_q+\beta_m+\epsilon_{mqt}$, in percentage points; positive
  values indicate higher Calculation credit than the average included model
  after absorbing question identity.}
  \label{tab:rd-clean-question-fe}
  \centering
  \small
  \begin{tabular}{lrrr}
    \toprule
    Model & RD-clean records & Raw calc & Question-FE effect \\
    \midrule
    Claude Opus 4.7 & $440$ & $87.6\%$ & $+2.8$ pp \\
    GPT-5.5 & $368$ & $86.3\%$ & $+1.4$ pp \\
    Claude Sonnet 4.6 & $444$ & $86.4\%$ & $-0.1$ pp \\
    GLM 5.1 & $395$ & $86.3\%$ & $+0.5$ pp \\
    Qwen 3.6 27B & $281$ & $86.2\%$ & $-0.1$ pp \\
    Kimi K2.6 & $332$ & $84.8\%$ & $-0.8$ pp \\
    Gemini 3 Flash & $282$ & $85.3\%$ & $-0.8$ pp \\
    Gemini 3.1 Pro & $181$ & $84.1\%$ & $-2.3$ pp \\
    Gemma 4 31B & $186$ & $84.1\%$ & $-0.6$ pp \\
    \bottomrule
  \end{tabular}
\end{table}

\section{Dataset Summary Statistics}
\label{app:dataset-stats}

\begin{table}[h]
  \caption{\textsc{BigFinanceBench} dataset summary statistics, computed over the $928$-item master
  release.}
  \label{tab:dataset-stats}
  \centering
  \small
  \begin{tabular}{lr}
    \toprule
    Statistic & Value \\
    \midrule
    Questions (total) & $928$ \\
    Distinct annotators & $52$ \\
    Distinct reviewers & $12$ \\
    \midrule
    Query length, mean / median / max chars & $324$ / $281$ / $1{,}479$ \\
    Answer length, median / p$75$ / p$95$ / p$99$ chars & $9$ / $34$ / $150$ / $384$ \\
    Answers $\le 20$ chars & $648$ ($70\%$) \\
    Answers $\ge 200$ chars & $33$ ($3.6\%$) \\
    \midrule
    Rubric criteria per question, mean / median / max & $16.9$ / $13$ / $192$ \\
    Rubric criteria (total) & $15{,}656$ \\
    Rubric criteria with explicit \texttt{[+N]} weight & $99.9\%$ \\
    Rubric points per question, mean / max & $39.1$ / $472$ \\
    Rubric points (total) & $36{,}241$ \\
    \midrule
    Source mentions: $10$-K & $495$ \\
    Source mentions: $10$-Q & $303$ \\
    Source mentions: press release / IR & $153$ \\
    Source mentions: other / market data & $148$ \\
    Source mentions: $8$-K & $108$ \\
    Source mentions: proxy / S-1 & $52$ \\
    Source mentions: web & $21$ \\
    Multi-source questions & $280$ ($30\%$) \\
    \bottomrule
  \end{tabular}
\end{table}

\begin{figure}[h]
  \centering
  \includegraphics[width=\linewidth]{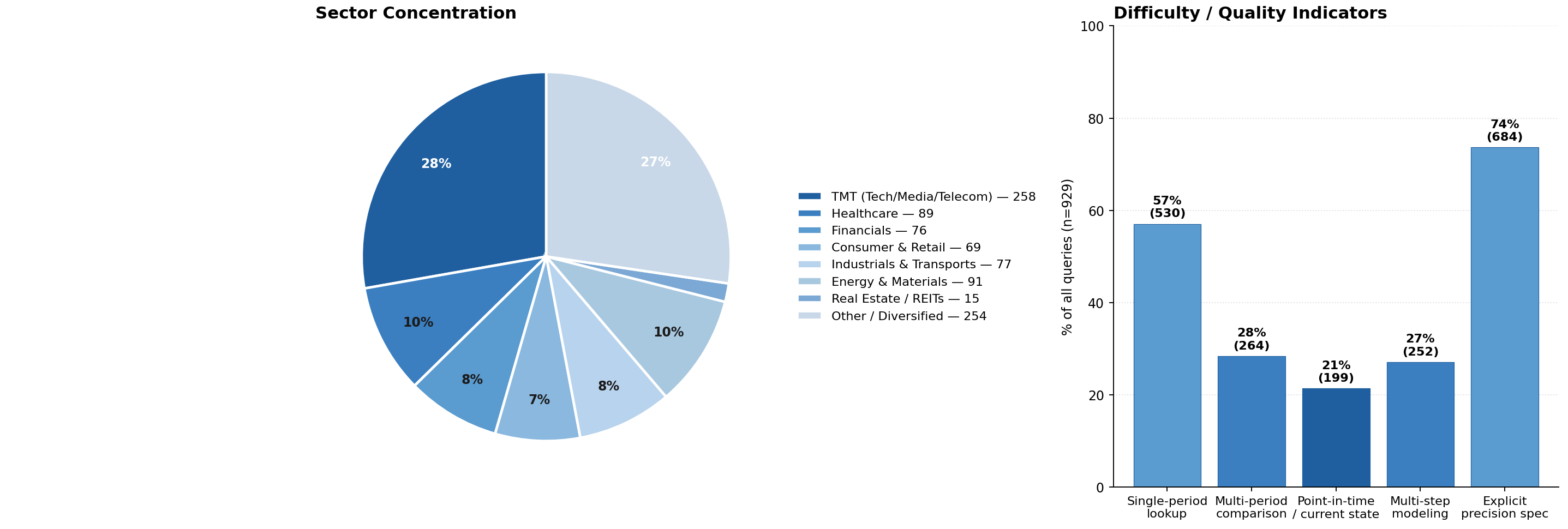}
  \caption{Coverage and difficulty composition of \textsc{BigFinanceBench}. (a) Sector
  concentration across the $928$ items. (b) Per-question difficulty / quality indicators; each
  item can carry several.}
  \label{fig:dataset-sector}
\end{figure}

Figure~\ref{fig:dataset-sector} gives the corresponding appendix marginals: sector coverage is
broad but concentrated in TMT and diversified items, while most questions specify precision
and many remain single-period lookups despite having compact final answers.

\section{Workflow / Skill Classification}
\label{app:workflow-classification}
The workflow and skill axes used by Figure~\ref{fig:dataset-workflow} are not authored fields
of the dataset; they are reproduced from the question text, reference answer, and rubric by a
deterministic LLM classifier. This appendix documents the classifier so the figure is
auditable end-to-end.
\paragraph{Taxonomy.} The classifier assigns each question exactly one of nine
workflow / decision types and exactly one of six analytical skills. The taxonomy strings are
pinned in \texttt{paper/code/classify\_question\_workflow\_v2.py} and re-imported by the plot
script, so the axis labels in Figure~\ref{fig:dataset-workflow} are guaranteed to match the
enum the classifier was constrained to. The nine workflows are \emph{Earnings Quality \& FSA},
\emph{Valuation \& Multiples}, \emph{M\&A \& Special Situations},
\emph{Capital Structure \& Returns}, \emph{KPIs \& Unit Economics},
\emph{Scenarios \& Forecasting}, \emph{Private Capital \& Buyside},
\emph{Governance \& Compensation}, and \emph{Capital Markets \& Trading};
the six skills are \emph{Single-Filing Lookup},
\emph{Cross-Filing Synthesis}, \emph{Accounting Adjustment}, \emph{Quantitative Modeling},
\emph{Forward / Scenario}, and \emph{External / Market Data}. A one-paragraph definition is
supplied to the classifier for each bucket; the full prompt is emitted verbatim by the
classifier script.
\paragraph{Model and decoding.} The classifier is GPT-5.5 with strict JSON-schema
structured output: the workflow and skill enums are the only legal values for the two
response fields, so off-taxonomy strings are physically unreturnable. We use the model's
default temperature so the structured-output path is honoured uniformly across runs. The
system prompt, user prompt, and per-bucket definitions are pinned to a single
\texttt{PROMPT\_VERSION} string that is stamped into the output CSV; mixing prompt versions
on resume is refused with an actionable error.
\paragraph{Three-sample voting.} Each question is classified three times. The three samples
share the same model and temperature, but the order in which the workflow and skill
enum values are listed inside the user prompt is shuffled deterministically per
(question id, sample index) using SHA-$256$-derived seeds. Order shuffling is the practical
control for recency / primacy bias in prompt enumerations: at the model's default
temperature, identical prompts return identical labels, so genuine ambiguity in a borderline
question only surfaces if the question is sensitive to the position of the candidate labels.
The CSV records all three votes per axis and the modal share; the assigned label is the
majority vote, with first-sample tie-break.
\paragraph{Reliability.} The classifier produces a complete (workflow, skill) pair for all
$928$ questions in the released master. Across those $928$ questions, the three samples
agree unanimously on workflow for $724$ ($78.0\%$), reach a $2$-of-$3$ majority for $154$
($16.6\%$), and split evenly across all three votes for $50$ ($5.4\%$). The corresponding
skill numbers are $711$ unanimous ($76.6\%$), $161$ majority ($17.3\%$), and $56$ tied
($6.0\%$). Borderline questions are therefore concentrated in a small minority of the
dataset and are individually identifiable in the published
\texttt{question\_workflow\_labels.csv} via the \texttt{workflow\_agreement} and
\texttt{skill\_agreement} columns.
\paragraph{Reproducing the figure.} The code for reproduction will be released together with the paper.

\section{Rubric-Criterion Stage Classification}
\label{app:rubric-stage-classification}
\begin{figure}[h]
  \centering
  \includegraphics[width=\linewidth]{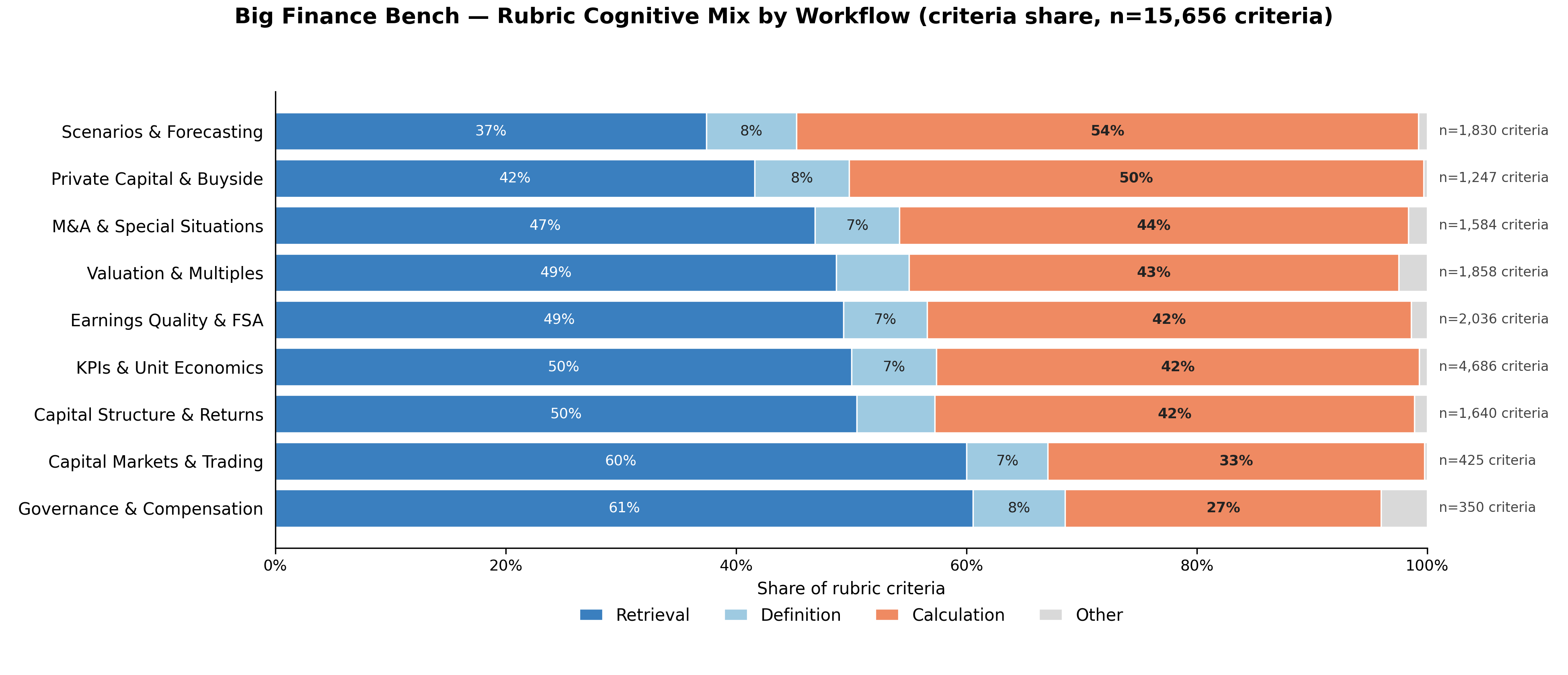}
  \caption{Rubric-criterion stage mix by workflow classified as \emph{Retrieval},
  \emph{Definition}, \emph{Calculation}, or \emph{Other}.}
  \label{fig:rubric-stage-by-workflow}
\end{figure}

Figures~\ref{fig:rubric-stage-by-workflow} and~\ref{fig:rubric-stage-accuracy} both condition
on a per-criterion \emph{stage} label. The labels are produced by a deterministic regex
classifier (\texttt{paper/code/classify\_rubric\_stages.py}); we deliberately did not use an
LLM here because the rubric prose is highly formulaic and a string-matching rule is auditable
criterion-by-criterion.
Across all rubric criteria, Retrieval accounts for $7{,}494$ lines ($47.9\%$), Definition for
$1{,}141$ ($7.3\%$), Calculation for $6{,}837$ ($43.7\%$), and Other for $184$ ($1.2\%$).
Point weighting shifts the mix toward Calculation: Retrieval accounts for $10{,}690$ points
($29.5\%$), Definition for $2{,}114$ ($5.8\%$), Calculation for $22{,}892$ ($63.2\%$), and
Other for $545$ ($1.5\%$). Figure~\ref{fig:rubric-stage-by-workflow} reports the corresponding
per-workflow stage mix.

\paragraph{Stages.} Each criterion is mapped to one of four stages. \emph{Retrieval} covers the
act of locating a specific value in a filing or source (typical leading verbs:
\textit{identifies}, \textit{records}, \textit{reports}, \textit{indicates}, \textit{notes},
\textit{states}). \emph{Definition} covers declaring a formula or methodology
(\textit{defines}, \textit{formula}, \textit{methodology}, \textit{equation},
\textit{assumes}). \emph{Calculation} covers arithmetic and quantitative transformations
(\textit{calculates}, \textit{computes}, \textit{sums}, \textit{adds}, \textit{subtracts},
\textit{multiplies}, \textit{divides}, \textit{adjusts}, \textit{normalises},
\textit{derives}, \textit{estimates}). \emph{Other} is the catch-all residual.
\paragraph{Matching procedure.} Each line is processed in two passes. First, a leading
adverb in the closed set $\{\textit{correctly}, \textit{properly}, \textit{accurately},
\textit{appropriately}, \textit{successfully}, \textit{clearly}, \textit{explicitly},
\textit{finally}\}$ is stripped so ``Correctly identifies \dots'' matches as
``identifies \dots''. A leading prepositional phrase of the form ``For \emph{noun~phrase},''
is also stripped so ``For CFO Jane Doe, reports adoption date \dots'' matches as
``reports adoption date \dots''. This handles a Governance \& Compensation authoring
convention that would otherwise leak the post-comma verb into the residual. Second, the first
whitespace-delimited token (lower-cased) is checked
against an ordered list of verb sets in the order \emph{Calculation}, \emph{Definition},
\emph{Retrieval}; the first set that contains the token wins. Author typos for
high-frequency verbs (e.g. \textit{idenfities}, \textit{identifes}, \textit{calcuates})
are listed alongside the
correctly-spelled forms so they do not leak into the residual. Criteria whose leading token is
in none of the verb sets land in \emph{Other}.
\paragraph{Distribution.} Across the $15{,}656$ rubric criteria in the released master, the
rules assign $47.9\%$ of \emph{criteria} to Retrieval, $7.3\%$ to Definition, $43.7\%$ to
Calculation, and $1.2\%$ to Other. Weighted by rubric \emph{points} the distribution shifts
substantially: Retrieval drops to $29.5\%$ of points while Calculation rises to $63.2\%$,
reflecting the authoring convention that conclusion-style calculation criteria carry the
largest \texttt{[+N]} weights. The Other residual at $1.2\%$ of criteria is dominated by
leading words that appear once or twice in the dataset (e.g. \textit{utilizes},
\textit{includes}, \textit{excludes}); none of the residual leading words exceed $0.13\%$ of
criteria individually, so widening the verb sets to absorb them would not change either
figure's per-stage shares by $0.1$~pp.
\paragraph{Reproduction.} \texttt{python code/classify\_rubric\_stages.py} writes one row per
\texttt{(qid, line\_idx)} to \texttt{paper/data/rubric\_line\_stages.csv} (committed alongside
the rest of the post-processed CSVs). The classifier is content-based and has no random
state; re-running on the same dataset is byte-identical.

\section{Harness Diagnostics}
\label{app:harness-diagnostics}

\begin{figure}[h]
  \centering
  \includegraphics[width=\linewidth]{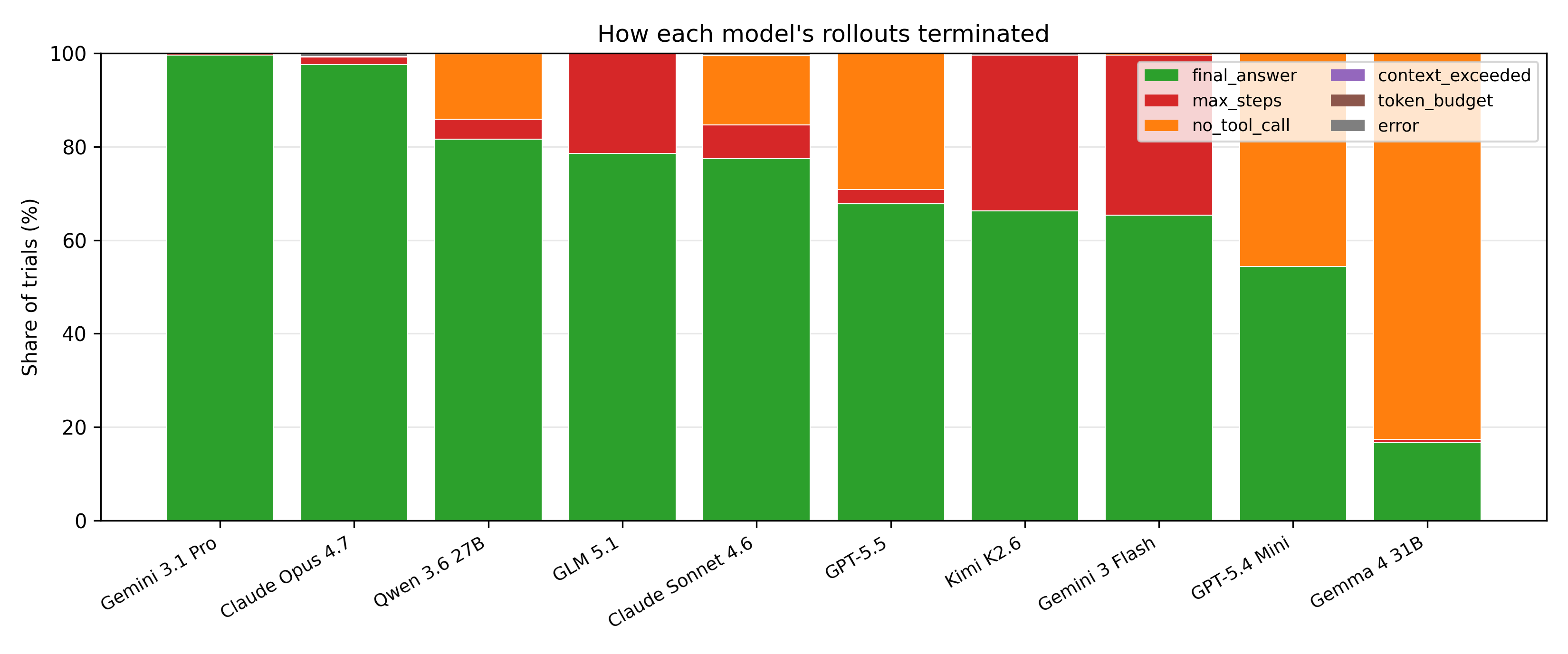}
  \caption{ReAct rollout stop-reason distribution per model.
  \texttt{final\_answer} indicates the model called the terminal tool;
  \texttt{max\_steps} that it exhausted the $50$-step budget;
  \texttt{no\_tool\_call} that it produced text without invoking any tool.
  Models are sorted by \texttt{final\_answer} share descending.}
  \label{fig:stop-reasons}
\end{figure}

\paragraph{ReAct stop-reason distribution.}
\label{app:stop-reasons}
Figure~\ref{fig:stop-reasons} breaks down the share of trials terminating in
each ReAct stop state: \texttt{final\_answer} (the model called the terminal
tool), \texttt{max\_steps} (it exhausted the $50$-step budget without
submitting), or \texttt{no\_tool\_call} (it produced text without calling any
tool, so the harness recorded the assistant text as the answer). Two
qualitatively distinct failure modes emerge.
\emph{Looping}, defined by a high \texttt{max\_steps} share, accounts for $33\%$ of Kimi
K2.6 trials, $34\%$ of Gemini 3 Flash trials, and $21\%$ of GLM~5.1 trials;
these systems often run out of budget without converging on an answer.
\emph{Skipping the workflow}, defined by a high \texttt{no\_tool\_call} share, is the
dominant Gemma 4~31B failure ($83\%$ of trials) and the dominant GPT-5.4~Mini
failure ($46\%$ of trials), and explains why those systems' final-answer
accuracy in Table~\ref{tab:benchmark-results} is substantially below their
rubric score: when they do submit an answer, it is generated from prior
knowledge rather than from the workflow the rubric is designed to check.

\begin{figure}[h]
  \centering
  \includegraphics[width=\linewidth]{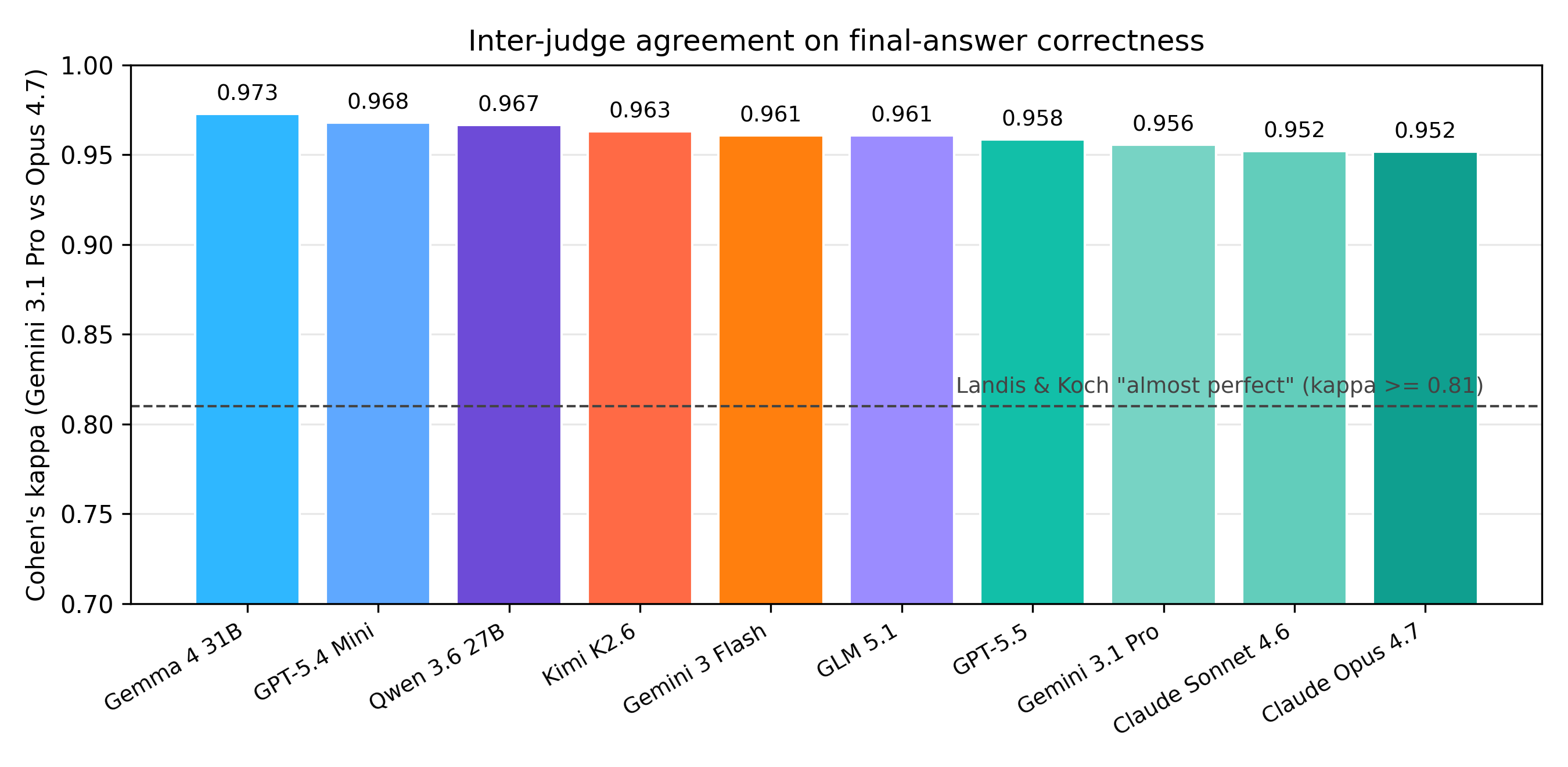}
  \caption{Per-model Cohen's $\kappa$ between the two judges (Gemini~3.1~Pro
  Preview and Claude Opus 4.7) on final-answer correctness, paired by
  $(\text{question}, \text{trial})$. All values exceed the $0.81$
  ``almost perfect'' threshold of \citet{landis1977measurement}.}
  \label{fig:inter-judge-kappa}
\end{figure}

\paragraph{Inter-judge agreement.}
\label{app:inter-judge-kappa}
The headline metrics in Table~\ref{tab:benchmark-results} are the mean across
two independent LLM judges (Gemini~3.1~Pro Preview and Claude Opus~4.7).
Figure~\ref{fig:inter-judge-kappa} reports Cohen's $\kappa$ on the binary
final-answer-correctness verdict between the two single-judge graders as a
robustness check on the averaging. Per-model $\kappa$ ranges from $0.952$
(Claude Opus~4.7 and Claude Sonnet~4.6, which tie at the floor) to $0.973$
(Gemma~4~31B), all comfortably above the $0.81$
``almost perfect'' threshold of \citet{landis1977measurement}.
On final-answer accuracy the two judges differ by under $1.3$ percentage
points on every model. On rubric score the disagreement is larger and
asymmetric: averaged over $(\text{question}, \text{trial})$ pairs, the Opus
judge marks $+1.9$~pp more credit than Gemini for Claude Opus~4.7 and $+0.9$~pp
for Claude Sonnet~4.6, but $+5.3$~pp for GPT-5.5, which is enough for GPT-5.5 to rank
third on rubric under the Gemini judge alone and first under the Opus judge
alone. The mean of both judges, which is what the headline metric reports, is
the natural way to be robust to that single-judge sensitivity.

\begin{figure}[h]
  \centering
  \includegraphics[width=\linewidth]{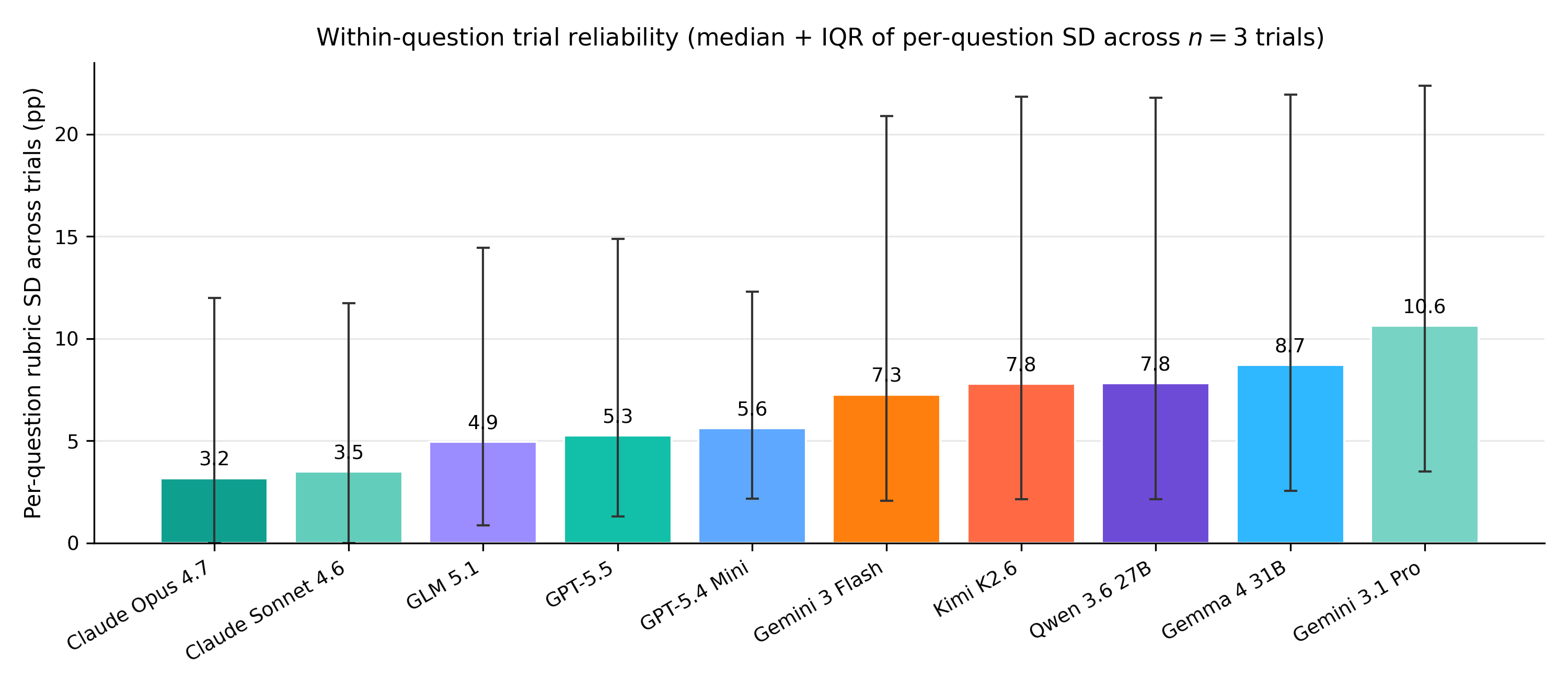}
  \caption{Per-question rubric-score standard deviation across the $n=3$
  trials per question (two-judge mean per trial). Bars are medians across
  questions; error bars span the $25$th to $75$th percentiles. Models are
  sorted by ascending median.}
  \label{fig:trial-reliability}
\end{figure}

\paragraph{Within-question trial reliability.}
\label{app:trial-reliability}
Inter-judge $\kappa$ controls for judge noise on a single trial; the
complementary axis is rollout noise across the $n=3$ trials per question for a
fixed judge pair. Figure~\ref{fig:trial-reliability} reports the median
per-question rubric SD across trials, with the interquartile range as error
bars, sorted by ascending median. The two Anthropic systems are most consistent
($3.2$~pp Opus, $3.5$~pp Sonnet); GLM~5.1, GPT-5.5, and GPT-5.4~Mini cluster
around $5$~pp; the open-weight middle and Gemini 3 Flash sit between $7.3$ and $7.8$~pp;
Gemini~3.1~Pro is the noisiest at $10.6$~pp. After the $n=3$ trial averaging
and macro-averaging across the $928$-question benchmark, the standard error
contributed by this rollout noise falls below $0.25$~pp for every model, so
the headline-metric ordering in Table~\ref{tab:benchmark-results} reflects
between-model differences rather than per-question rollout variance.

\section{Conditional Drill-Downs}
\label{app:conditional-drilldowns}

\begin{figure}[h]
  \centering
  \includegraphics[width=\linewidth]{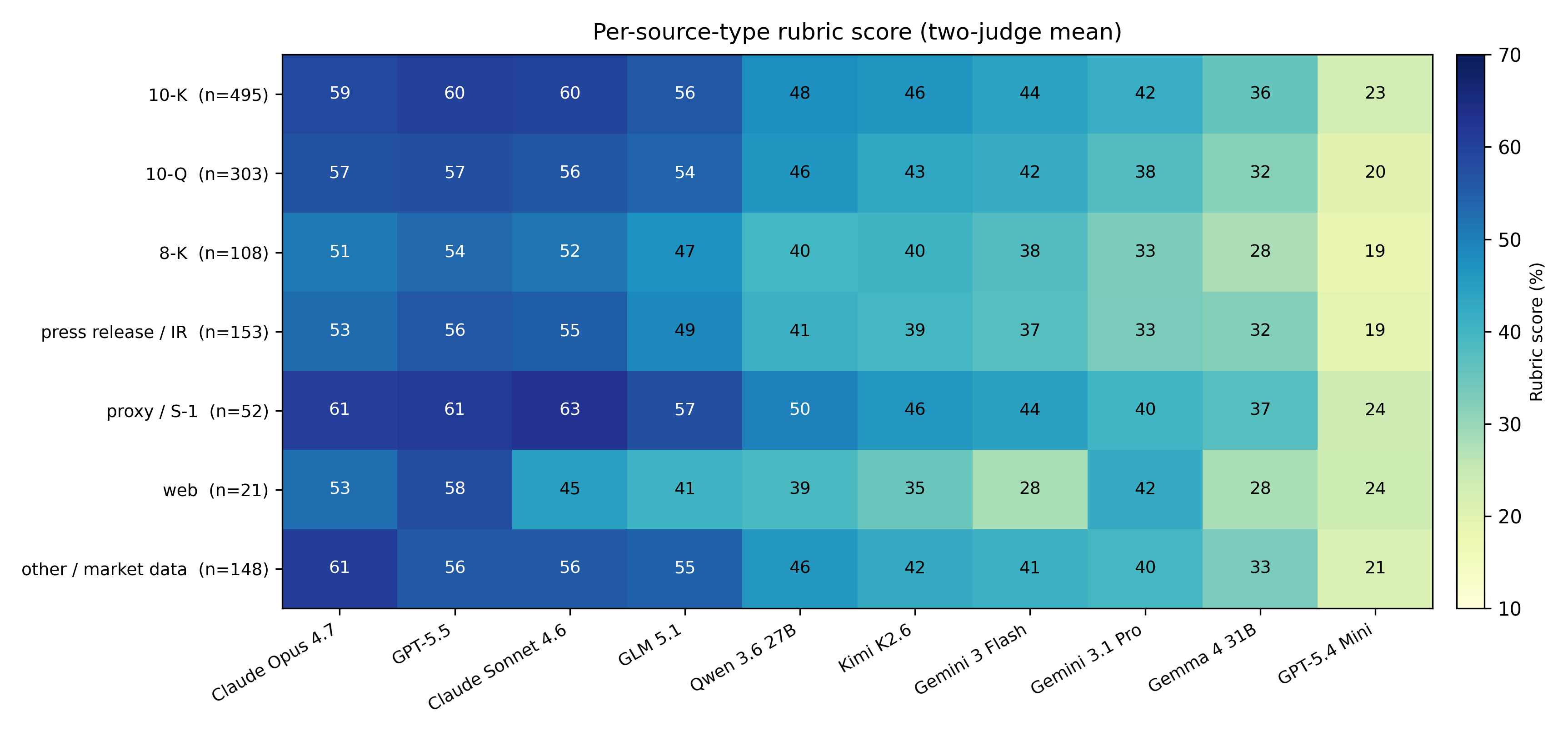}
  \caption{Per-model rubric score conditioned on the source-document type the
  question requires. Source buckets follow the appendix taxonomy; questions
  can appear in multiple buckets. Models are sorted left-to-right by overall
  rubric score; cell values are percentages.}
  \label{fig:source-type-accuracy}
\end{figure}

\paragraph{Source-document conditioning.}
\label{app:source-type}
Different question types are grounded in different SEC filing types (10-K,
10-Q, 8-K, proxy/S-1) or non-SEC sources (press releases, IR pages, web
pages, market data). Figure~\ref{fig:source-type-accuracy} conditions the
rubric score on the source bucket the question requires (questions that
require multiple source types appear in each bucket they cite). Model rank
ordering is largely preserved across buckets: the three top systems (Claude
Opus~4.7, GPT-5.5, Claude Sonnet~4.6) are within $3$ percentage points of
each other in each of the four largest buckets (10-K, 10-Q, 8-K, press
release / IR), and within each model the per-bucket rubric across those
four buckets varies by under $8$ percentage points. The one consistent gap
is between 10-K- and 8-K-grounded questions: 8-K rubric scores sit $6$ to $9$
percentage points below 10-K scores for every evaluated model in the top
half of the table (Opus $58 \to 51$, GPT-5.5 $60 \to 54$, Sonnet $59 \to 52$,
GLM~5.1 $56 \to 47$, Qwen 3.6 27B $48 \to 39$), consistent with 8-K events
being less structurally standardised than periodic filings. Smaller buckets
(proxy/S-1, web) are not large enough to support firm conclusions.

\begin{figure}[h]
  \centering
  \includegraphics[width=\linewidth]{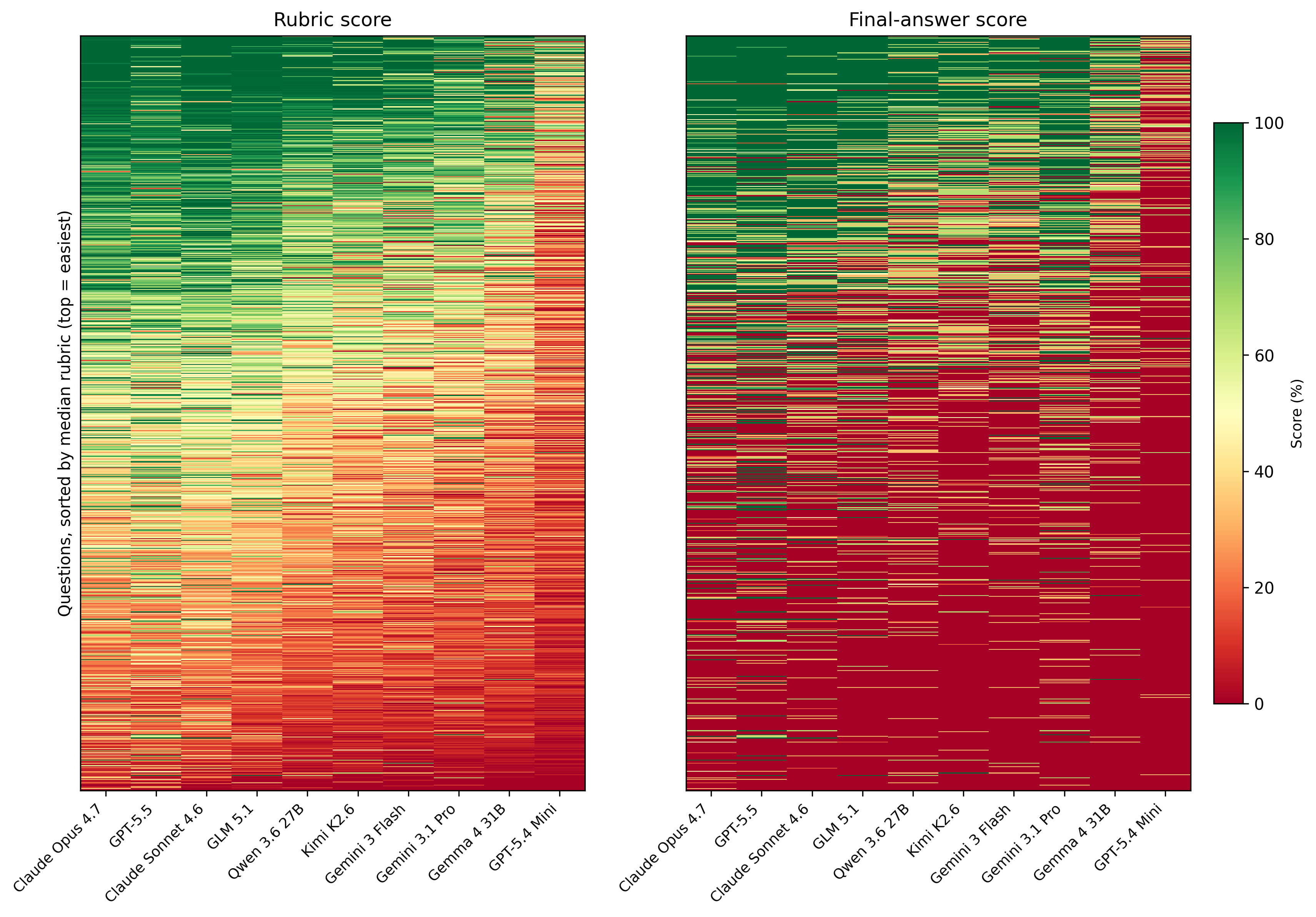}
  \caption{Per-(question, model) score, rubric (left) vs final-answer (right). Each row is one
  question; rows are shared between panels and sorted by per-question median rubric (top =
  easiest). Each column is one model, sorted by overall rubric. Cell colour is the
  per-(question, model) two-judge mean score across $n=3$ trials; final-answer scores live in
  $\{0, \tfrac{1}{6}, \tfrac{2}{6}, \dots, 1\}$ and rubric scores are continuous in $[0, 1]$.}
  \label{fig:difficulty-distribution}
\end{figure}

\paragraph{Per-question difficulty structure.}
\label{app:difficulty-distribution}
Figure~\ref{fig:difficulty-distribution} shows the full question-by-model score surface under
rubric and final-answer scoring.

\section{Routing-amenable Frontier}
\label{app:routing-ladder}

\begin{figure}[h]
  \centering
  \includegraphics[width=\linewidth]{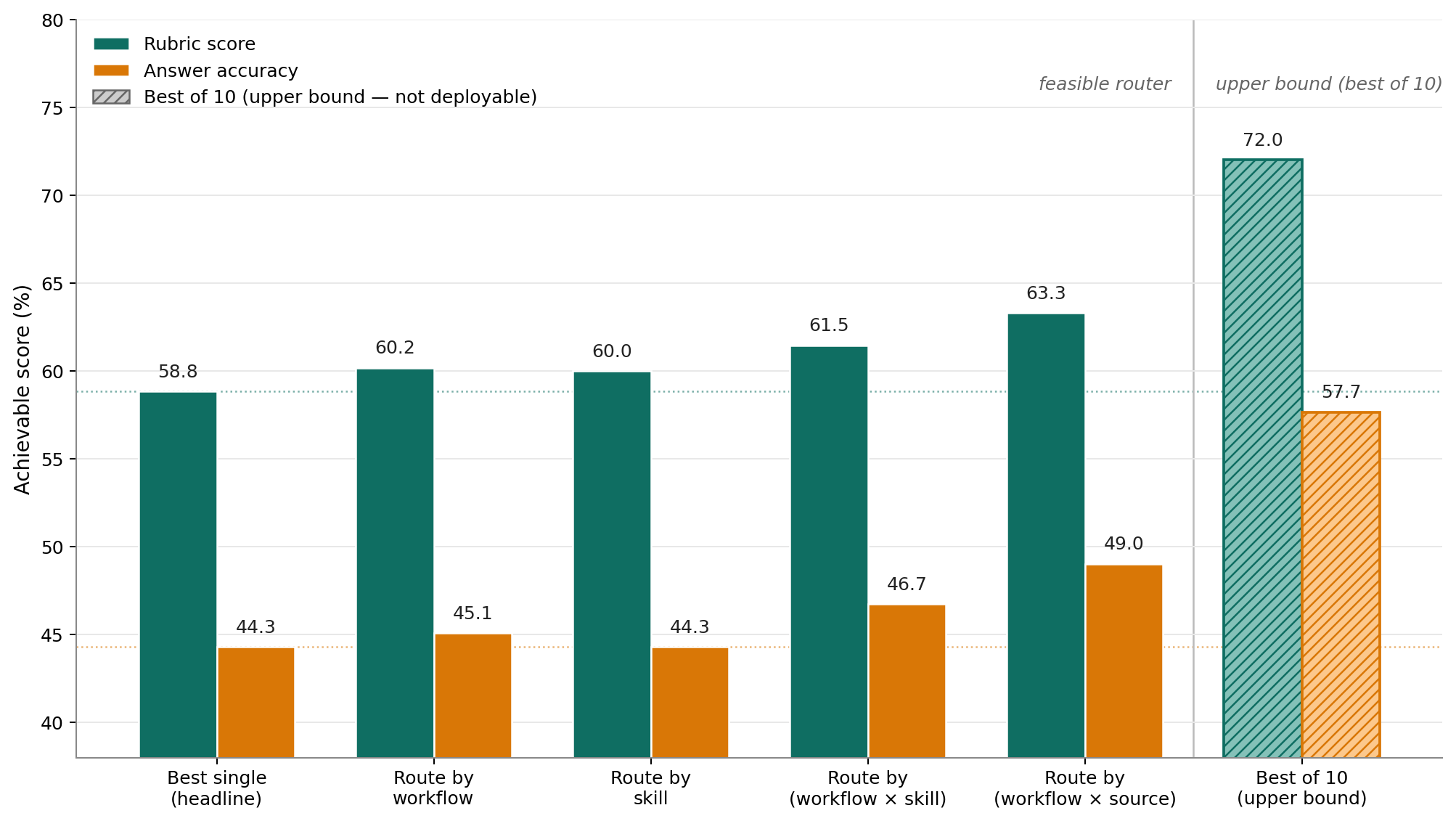}
  \caption{Routing ladder: achievable rubric (teal) and final-answer accuracy (orange) at
  increasing routing granularity, computed across the ten evaluated models. Solid bars are
  deployable routers using observable question features; the hatched rightmost rung is a
  best-of-10 upper bound that requires ground-truth selection and is therefore not deployable.
  Dotted reference lines mark headline best-single performance.}
  \label{fig:routing-ladder}
\end{figure}

Figure~\ref{fig:routing-ladder} shows achievable rubric and final-answer accuracy at
increasing routing granularity, computed across the ten evaluated models. Solid bars are
deployable routers that select a model from observable question features (workflow, skill,
source bucket); the hatched rightmost rung is a best-of-10 upper bound that selects, per
question, the model with the highest score on the metric in question (rubric-best for the
rubric number, answer-best for the answer-accuracy number). The upper bound is not a
deployable system since it requires the metric grade itself to identify the per-question
winner. Coarse (workflow $\times$ source)
routing buys $+4.5$~pp rubric / $+4.7$~pp answer accuracy over the best single model; the
best-of-10 upper bound is $+13.2$~pp / $+13.4$~pp. Both metrics agree that the achievable
frontier on \textsc{BigFinanceBench} is meaningfully wider than any single evaluated model
contributes, and that the gap between feasible coarse routers and the best-of-10 upper bound
($\sim$$9$~pp on each metric) is the headroom available to a learned router that conditions
on question features.

\begin{figure}[h]
  \centering
  \includegraphics[width=\linewidth]{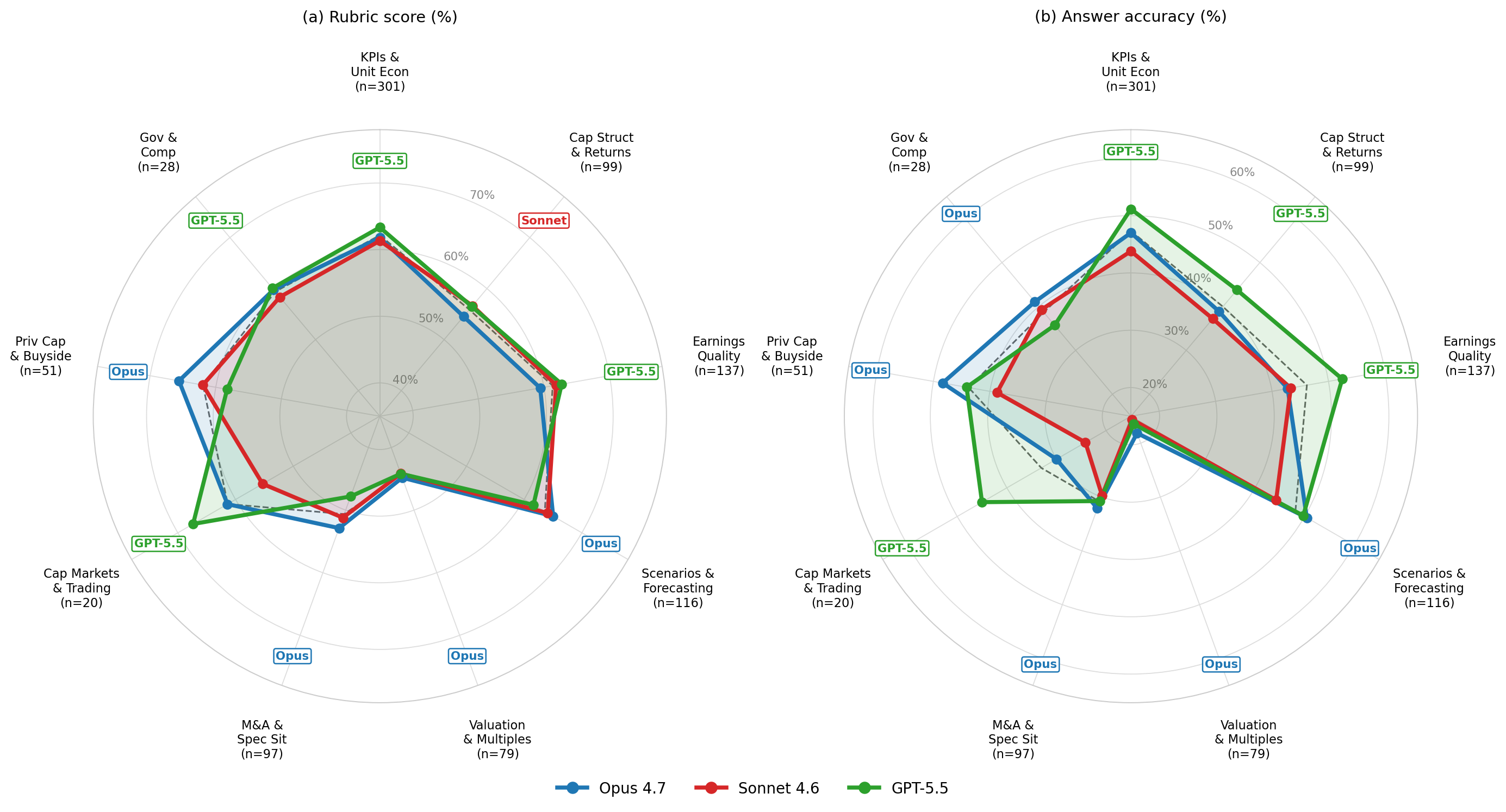}
  \caption{Per-workflow rubric (left) and final-answer accuracy (right) of the top-3 closed
  models in absolute form, complementing Figure~\ref{fig:partition-radar}. Boxed labels mark
  the workflow leader; the dashed gray polygon traces the per-workflow $3$-model mean (the
  reference that Figure~\ref{fig:partition-radar} subtracts to produce its centered values).
  Polygons cluster tightly at this scale, which is the visual hazard the centering used in
  the main text avoids.}
  \label{fig:partition-radar-absolute}
\end{figure}

Figure~\ref{fig:partition-radar-absolute} replots the per-workflow polygons of
Figure~\ref{fig:partition-radar} in absolute mean rubric and final-answer accuracy. Polygons
cluster tightly at the absolute scale because the across-workflow variation (some workflows
are inherently easier) dwarfs the across-model variation; the centering used in the main-text
figure removes that shared workflow-difficulty signal. Concretely, for each workflow and
metric we first compute each top-3 model's absolute mean score, then subtract the mean across
the three models for that same workflow. The main-text radar therefore reports percentage
points relative to the per-workflow $3$-model mean, not standard deviations. For plotting
only, these signed deviations are shifted by a constant positive radius on a zoomed
percentage-point scale; the radial tick labels report the original signed deviations.

\clearpage

\end{document}